\documentclass[10pt,twocolumn,letterpaper]{article}

\usepackage[pagenumbers]{cvpr} % To force page numbers, e.g. for an arXiv version

\definecolor{cvprblue}{rgb}{0.21,0.49,0.74}
\usepackage[pagebackref,breaklinks,colorlinks,allcolors=cvprblue]{hyperref}
\usepackage{graphicx}
\usepackage{amsmath}
\usepackage{amsthm}
\usepackage{makecell}
\usepackage{longtable}
\usepackage{multirow}
\usepackage{multicol}
\usepackage{lineno}
\usepackage{xcolor}
\usepackage{booktabs}
\usepackage{caption}
\usepackage{epsfig}
\usepackage{amssymb}
\usepackage{pifont}
\usepackage{epsfig}
\usepackage{afterpage}
\usepackage{amsmath} 
\usepackage{cuted}
\usepackage{float}
\usepackage{bbding}
\usepackage[table]{xcolor}  
\usepackage{bm}
\usepackage{setspace}

\def\paperID{} % *** Enter the Paper ID here
\def\confName{CVPR}
\def\confYear{2026}

\title{EgoSound: Benchmarking Sound Understanding in Egocentric Videos}

\author{
    Bingwen Zhu$^{1,2*}$\qquad
    Yuqian Fu$^{1,3,6*\dagger}$\qquad
    Qiaole Dong$^{1}$\qquad
    Guolei Sun$^{4}$\qquad 
    Tianwen Qian$^{5}$\qquad \\
    Yuzheng Wu$^{1}$\qquad
    Danda Pani Paudel$^{3}$\qquad
    Xiangyang Xue$^{1}$\qquad  
    Yanwei Fu$^{1,2}$\qquad \\
    $^{1}$Fudan University\qquad
    $^{2}$Shanghai Innovation Institute\qquad
    $^{3}$INSAIT, Sofia University  “St. Kliment Ohridski”\qquad \\
    $^{4}$Nankai University\qquad
    $^{5}$East China Normal University\qquad
    $^{6}$KAUST
}

\begin{document}
\maketitle
\footnotetext[1]{Equal Contribution.}      % 1 -> *
\footnotetext[2]{Corresponding Author.}    % 2 -> †
\begin{strip}
\centering
\vspace{-4em}
\includegraphics[width=1.0\textwidth]{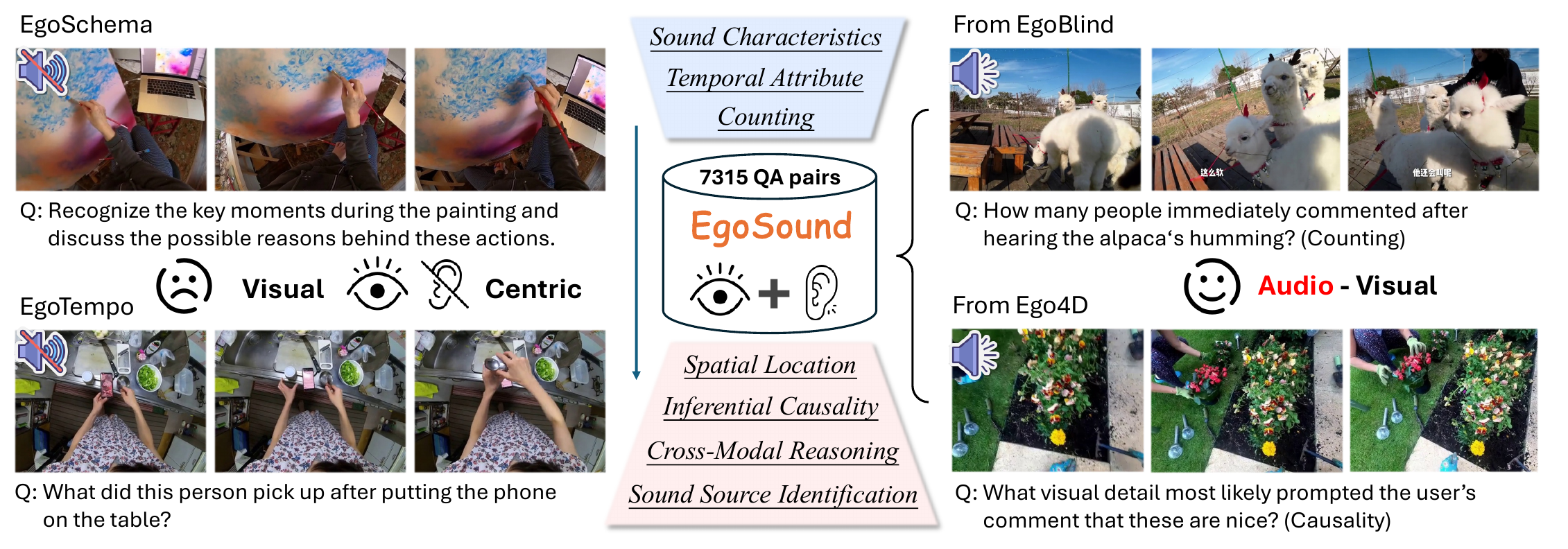} 
\vspace{-2em}
\captionof{figure}{
\textbf{EgoSound \textit{vs} existing egocentric Video Question Answering (VideoQA).} Prior datasets (left)~\cite{mangalam2023egoschema,plizzari2025omnia} focus solely on vision-centric question answering with no awareness of audio, whereas EgoSound (right) constructs a more complex and comprehensive audio-visual QA dataset tailored for sound understanding. It is built from two dataset sources~\cite{xiao2025egoblind,grauman2022ego4d}, includes 900 videos and 7315 high-quality QA pairs, and spans seven task categories—making it a benchmark that can both listen and see.}
\label{fig:teaser}
\end{strip}

\begin{abstract}
Multimodal Large Language Models (MLLMs) have recently achieved remarkable progress in vision-language understanding. Yet, human perception is inherently multisensory, integrating sight, sound, and motion to reason about the world. Among these modalities, sound provides indispensable cues about spatial layout, off-screen events, and causal interactions, particularly in egocentric settings where auditory and visual signals are tightly coupled. To this end, we introduce EgoSound, the first benchmark designed to systematically evaluate egocentric sound understanding in MLLMs. EgoSound unifies data from Ego4D and EgoBlind, encompassing both sighted and sound-dependent experiences. It defines a seven-task taxonomy spanning intrinsic sound perception, spatial localization, causal inference, and cross-modal reasoning. Constructed through a multi-stage auto-generative pipeline, EgoSound contains 7315 validated QA pairs across 900 videos. Comprehensive experiments on nine state-of-the-art MLLMs reveal that current models exhibit emerging auditory reasoning abilities but remain limited in fine-grained spatial and causal understanding. EgoSound establishes a challenging foundation for advancing multisensory egocentric intelligence, bridging the gap between seeing and truly hearing the world. 
% Dataset and codes are available at \href{https://github.com/groolegend/EgoSound}{https://github.com/groolegend/EgoSound}.
Project page: \href{https://groolegend.github.io/EgoSound/}{https://groolegend.github.io/EgoSound/}.
\end{abstract}    
\section{Introduction}
\label{sec:intro}
Multimodal Large Language Models (MLLMs) have recently demonstrated remarkable progress in integrating vision and language, enabling sophisticated visual understanding and reasoning. Yet, true human-like perception extends far beyond vision—it is inherently multisensory, grounded in the seamless integration of sound, touch, and motion. Among these, sound plays a particularly vital role: it conveys spatial cues, reveals off-screen events, and encodes the causality and intent behind interactions. This becomes especially critical in egocentric settings, where the auditory and visual streams are deeply intertwined, capturing the world as directly experienced by the wearer.

Despite this importance, research in egocentric perception has remained overwhelmingly visual-centric. Most prior works~\cite{plizzari2025omnia,li2025egocross,fu2025object,zhang2025egonight,rodin2025easg,xiao2025egoblind,goletto2024amego,jia2022egotaskqa,cheng2024egothink,Fan_2019_ICCV,huang2024egoexolearn,ye2024mm,zhou2025egotextvqa,majumdar2024openeqa,wen2025ai,li2025clivis,mahdi2025exo2egosyn,fu2025cross,pan2025v} focus on recognizing and predicting events visible in the scene, while treating audio as secondary—or ignoring it entirely. Perceiving the world egocentrically without sound is akin to navigating a silent world, fundamentally limiting the depth of understanding. For instance, the sharp hiss of steam or the sudden clatter of metal carries vital information that vision alone may miss. Likewise, for individuals with visual impairments, sound is not auxiliary but essential for navigation and situational awareness.

To fill this gap, we introduce \textbf{EgoSound}, a new benchmark that systematically evaluates the egocentric sound understanding capabilities of MLLMs~\cite{cheng2024videollama,tang2025video,yao2024minicpm,xu2025qwen2,xu2025qwen3,comanici2025gemini}. EgoSound is the first dataset explicitly designed to study nuanced audio-visual reasoning from a first-person perspective, encompassing both environmental sounds from human-object interactions and human dialogues that drive contextual understanding. Its mission is simple yet profound: to enable models that can hear, not just see, from a first-person viewpoint.
A distinguishing feature of EgoSound is its multi-source design, which integrates videos from both the large-scale Ego4D dataset~\cite{grauman2022ego4d}, capturing a wide range of everyday activities, and the EgoBlind dataset~\cite{xiao2025egoblind}, focusing on scenarios where auditory perception is essential for understanding, interaction, and navigation. This combination provides a comprehensive coverage of egocentric experiences, spanning from visually guided to sound-dependent contexts.

We further propose a novel taxonomy of seven egocentric sound tasks that span both unimodal and multimodal reasoning—from intrinsic sound properties (e.g., Sound Characteristics, Counting, Temporal Attribute) to complex audio-visual reasoning (e.g., Spatial Location, Source Identification, Inferential Causality, Cross-Modal Reasoning). To construct EgoSound, we develop a multi-stage data curation pipeline leveraging modern generative models (Qwen2.5-VL~\cite{bai2025qwen2}, Gemini-2.5~\cite{comanici2025gemini}, GPT-4o~\cite{hurst2024gpt}). The pipeline first identifies key human-object interactions, then generates rich audio-centric captions, and finally constructs high-quality, open-ended question–answer pairs focused on sound-related reasoning. The final dataset comprises 7315 validated Q\&A pairs over 900 rigorously filtered videos, ensuring strong fidelity and task diversity.

We evaluate eight state-of-the-art MLLMs, including models from the Qwen-Omni~\cite{xu2025qwen2,xu2025qwen3}, video-SALMONN 2+~\cite{tang2025video}, VideoLLaMA2.1~\cite{cheng2024videollama}, and MiniCPM~\cite{yao2024minicpm} families, as well as the egocentric-specialized EgoGPT~\cite{yang2025egolife}. Our comprehensive experiments reveal that while current MLLMs exhibit emerging auditory reasoning abilities, they still struggle with fine-grained spatial, temporal, and causal inference based on sound. EgoSound thus establishes a challenging, high-quality benchmark for multisensory egocentric intelligence—bridging a critical gap in current MLLMs and paving the way for future research toward models that can listen, understand, and reason about the full multisensory world.

Our main contributions are summarized as follows:
\begin{itemize}
    \item \textbf{EgoSound:} The first large-scale benchmark for egocentric sound understanding in MLLMs, featuring data from both sighted (Ego4D) and blind (EgoBlind) perspectives.
    \item \textbf{A Novel Task Taxonomy:} Seven tasks covering intrinsic sound perception, spatial reasoning, causal inference, and cross-modal understanding.
    \item \textbf{High-Quality Dataset:} 7315 validated open-ended Q\&A pairs created via a rigorous, multi-stage curation pipeline centered on sound events.
    \item \textbf{Comprehensive Benchmarking:} Evaluation of nine cutting-edge MLLMs, uncovering key challenges in egocentric audio-visual reasoning and establishing strong baselines for future research.
\end{itemize}

\begin{table*}[h]
\centering
\resizebox{0.9\textwidth}{!}{%
\begin{tabular}{@{}l|ccccccc@{}}
    \toprule
    \textbf{Dataset} &  Video Length  &  Clips  & QA Pairs & Categories & Sound Questions  & Multiple Sources & Open-ended \\
    \midrule
    EgoVQA~\cite{Fan_2019_ICCV} & (25, 100)s & 520 & 0.6k & 5 &  \ding{55} & \ding{55} & \Checkmark \\
    EgoTaskQA~\cite{jia2022egotaskqa} & 25s & 2336 & 40k & 4 & \ding{55} & \ding{55} & \Checkmark  \\
    EgoSchema~\cite{mangalam2023egoschema} & 3min & 1981 & 5k & - & \ding{55} & \ding{55} & \ding{55}  \\
    EgoThink~\cite{cheng2024egothink} & - & - & 0.75k  & 6 & \ding{55} & \ding{55} & \Checkmark \\
    AMEGO~\cite{goletto2024amego} & 14min & 100 & 20.5k & 8 & \ding{55} & \ding{55} & \ding{55} \\
    EgoTempo~\cite{plizzari2025omnia} & 45s & 365 & 0.5k & 10 &  \ding{55} & \ding{55} & \Checkmark \\
    EASG-Bench~\cite{rodin2025easg} & 3.1min & 221 & 1.8k & 5 & \ding{55} & \ding{55} & \Checkmark \\
    EgoCross~\cite{li2025egocross} & 22.5s & 798 & 0.95k & 4 &  \ding{55} & \Checkmark & \Checkmark \\
    \midrule
    \cellcolor{cyan!10} \textbf{EgoSound} (ours)  & \cellcolor{cyan!10} 59s & \cellcolor{cyan!10}900 & \cellcolor{cyan!10}7.3k & \cellcolor{cyan!10}7 &  \cellcolor{cyan!10}\Checkmark &  \cellcolor{cyan!10}\Checkmark & \cellcolor{cyan!10}\Checkmark \\
    \bottomrule
  \end{tabular}}
  \vspace{-0.05in}
\caption{Comparison with existing egocentric video QA benchmarks. EgoSound is distinguished by its focus on sound-centric reasoning.}
  \vspace{-0.15in}
\label{tab:dataset}
\end{table*}

\section{Related Work}
\label{sec:related}
\paragraph{Egocentric Video Question Answering.}
To advance video question answering (VideoQA) on egocentric videos, numerous datasets~\cite{damen2018scaling, EPICSOUNDS2025, wang2023holoassist, grauman2024ego, xiao2025egoblind, perrett2025hd} and benchmarks~\cite{Fan_2019_ICCV,jia2022egotaskqa,mangalam2023egoschema,cheng2024egothink,goletto2024amego,plizzari2025omnia,rodin2025easg,li2025egocross, zhang2025egonight, yang2025egolife, xu2025tog} have been proposed. Early and representative ones include EgoVQA~\cite{Fan_2019_ICCV}, EgoTaskQA~\cite{jia2022egotaskqa}, and EgoSchema~\cite{mangalam2023egoschema}, each focusing on different aspects and characteristics of first-person understanding, covering a range of tasks such as descriptive, predictive, explanatory, and counterfactual reasoning. More recently, several complementary benchmarks have also been introduced, broadening the evaluation scope. For example, EgoThink~\cite{cheng2024egothink} assesses models’ ability to “think” from a first-person perspective across multiple reasoning dimensions; AMEGO~\cite{goletto2024amego} emphasizes long-term temporal reasoning and memory over extended egocentric videos; EgoTempo~\cite{plizzari2025omnia} focuses on temporal understanding through long-horizon questions; EASG-Bench~\cite{rodin2025easg} introduces scene-graph-based QA to capture spatio-temporal relations; EgoCross~\cite{li2025egocross} explores cross-domain generalization across diverse first-person scenarios such as surgery and sports;
Despite covering diverse visual and cognitive skills, most existing egocentric QA benchmarks rely solely on visual cues, overlooking the rich auditory context in first-person videos. In contrast, EgoSound focuses on auditory-(visual) cues to promote a more comprehensive understanding of egocentric scenes. Tab.~\ref{tab:dataset} provides a multidimensional comparison between EgoSound and prior egocentric QA benchmarks.

\paragraph{Audio(-Visual) Question Answering.}
Audio-based Question Answering (Audio-QA) aims to reason about spatial and semantic information from audio or audio–visual observations to answer natural language queries.  Early research on audio reasoning primarily focused on localization and detection rather than QA, such as STARSSS23~\cite{shimada2023starss23}.
%for audio–visual sound localization and detection. 
Subsequent work shifted toward question answering with explicit spatial reasoning benchmarks. SpatialSoundQA~\cite{zheng2024bat} and AQAPHY~\cite{wang2025teaching} introduced different benchmark for Audio-QA, with distinct focuses of reasoning tasks. SAVVY~\cite{chen2025savvy} and Magnet~\cite{chowdhury2025magnet} further extend to audio–visual QA, emphasizing the joint understanding of auditory and visual modalities.  
Despite these advances, Audio(-Visual) QA in the \emph{egocentric} domain remains largely unexplored, overlooking the fact that audio plays a crucial role in understanding and grounding the world from a first-person perspective. 
%Thus, in this paper, we are motivated to bridge this gap by establishing the first dedicated audio(-visual) question answering testbed for the egocentric domain.

\paragraph{Multimodal Large Language Models.}
Multimodal large language models (MLLMs) have advanced rapidly, with a surge of commercial systems (e.g., GPT-4~\cite{achiam2023gpt}, Gemini 2.5 Pro~\cite{comanici2025gemini}) and open-source counterparts (e.g., Qwen2.5-VL~\cite{bai2025qwen2}, InternVL~\cite{zhu2025internvl3}, Video-LLaMA~\cite{cheng2024videollama, zhang2025videollama}, MiniCPM~\cite{yao2024minicpm}, video-SALMONN~\cite{tang2025video}, and Qwen-Omni~\cite{xu2025qwen2, xu2025qwen3}),  which demonstrate strong reasoning ability across benchmarks including MLVU~\cite{zhou2024mlvu}, LVBench~\cite{wang2025lvbench}, and EgoSchema~\cite{mangalam2023egoschema}.
In the egocentric domain, concurrent progress in both models and datasets has led to several pretrained egocentric MLLMs, such as EgoVLPv2~\cite{pramanick2023egovlpv2} and xEgoGPT~\cite{yang2025egolife}, 
However, \textit{audio}, a key cue for human perception and situational awareness, remains underexplored in both model design and evaluation. 
%Most existing MLLMs emphasize visual recognition or text-based reasoning, leaving multisensory reasoning insufficiently examined. For example, even the world-wide leading models such as GPT-4~\cite{achiam2023gpt} doesn't support the reasoning on coupled  audio–visual video inputs. 
Therefore, in this work, we focus on evaluating how MLLMs perceive and reason over sound cues and joint audio–visual cues under a first-person perspective.

\section{EgoSound Dataset}
\label{sec:method}
We introduce EgoSound, a novel benchmark to systematically evaluating performance of MLLMs across a variety of egocentric sound understanding tasks. Egosound covers both sighted and blind perspectives, and consists of egocentric videos spanning various visual scenarios. In this section, we provide a comprehensive introduction
to the EgoSound benchmark. We begin by discussing the video data selection, and the taxonomy of
question-answering tasks, followed by an explanation of the data curation pipeline, and conclude with dataset statistics.

\begin{figure*}[t]
  \centering
  \includegraphics[width=0.9\linewidth]{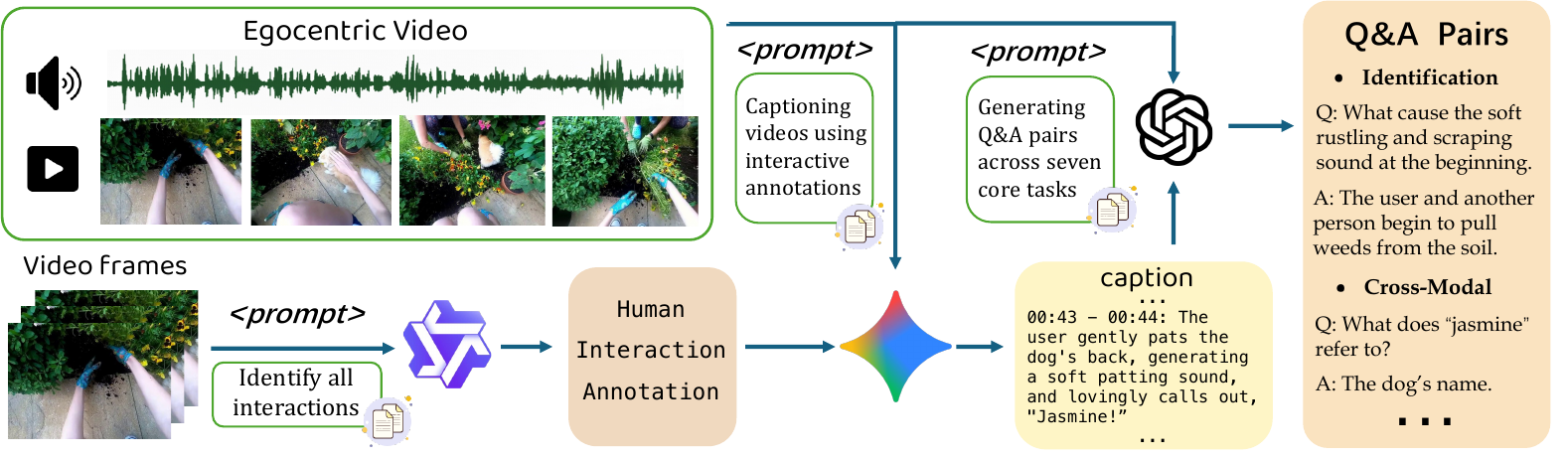}
  \caption{\textbf{Overview of the EgoSound data curation pipeline.} We first identifies human interaction events, then generates interaction-grounded and sound-centric audio-visual captions, and finally build visually-verified OpenQA pairs corresponding to the seven core tasks.}
  \label{fig:pipeline}
  \vspace{-0.1in}
\end{figure*}

\subsection{Data Collection and Filtering}
The videos included in EgoSound are carefully sourced from egocentric datasets with rich audio, including Ego4D~\cite{grauman2022ego4d}, and EgoBlind~\cite{xiao2025egoblind}. This selection ensures a broad spectrum of acoustic environments and contexts. 
Ego4D~\cite{grauman2022ego4d}, as the most extensive egocentric video dataset, contributes a wide variety of scenarios from daily life and work. The inclusion of EgoBlind~\cite{xiao2025egoblind} is particularly unique, offering data recorded by blind individuals where auditory cues are often central to daily navigation and interaction. Collectively, these sources ensure that our dataset covers a wide range of human activities such as sports, learning, doing chores, cooking, commuting, driving, shopping, playing instruments, and filming, across numerous indoor and outdoor environments. The videos in EgoSound span a broad range of durations, from short clips of 5 seconds to extended recordings of up to 5 minutes, capturing both brief atomic sounds and more complex, temporally evolving acoustic events.

To ensure data quality, the construction of EgoSound involved a rigorous filtering process for both audio and visual streams.
For the audio, we first discarded videos with extended silence, excessive background noise, or unintelligible speech. The remaining clips were then carefully trimmed to retain segments rich in meaningful sound events. This process concentrates the data on high-quality sound suitable for generating audio-centric question-answer pairs.
Visually, we removed clips that were static or monotonous. We specifically retained segments that feature dynamic human activities and rich object interactions.
Through this dual filtering strategy, we curated 900 egocentric videos featuring rich and complex audio-visual scenarios. The filtering process is crucial for creating challenging and high-quality queries to effectively evaluate the audio-visual understanding capabilities of MLLMs.

\subsection{Task Taxonomy}
\label{subsec:task}
\begin{figure*}[!ht]
  \centering
  \includegraphics[width=\linewidth]{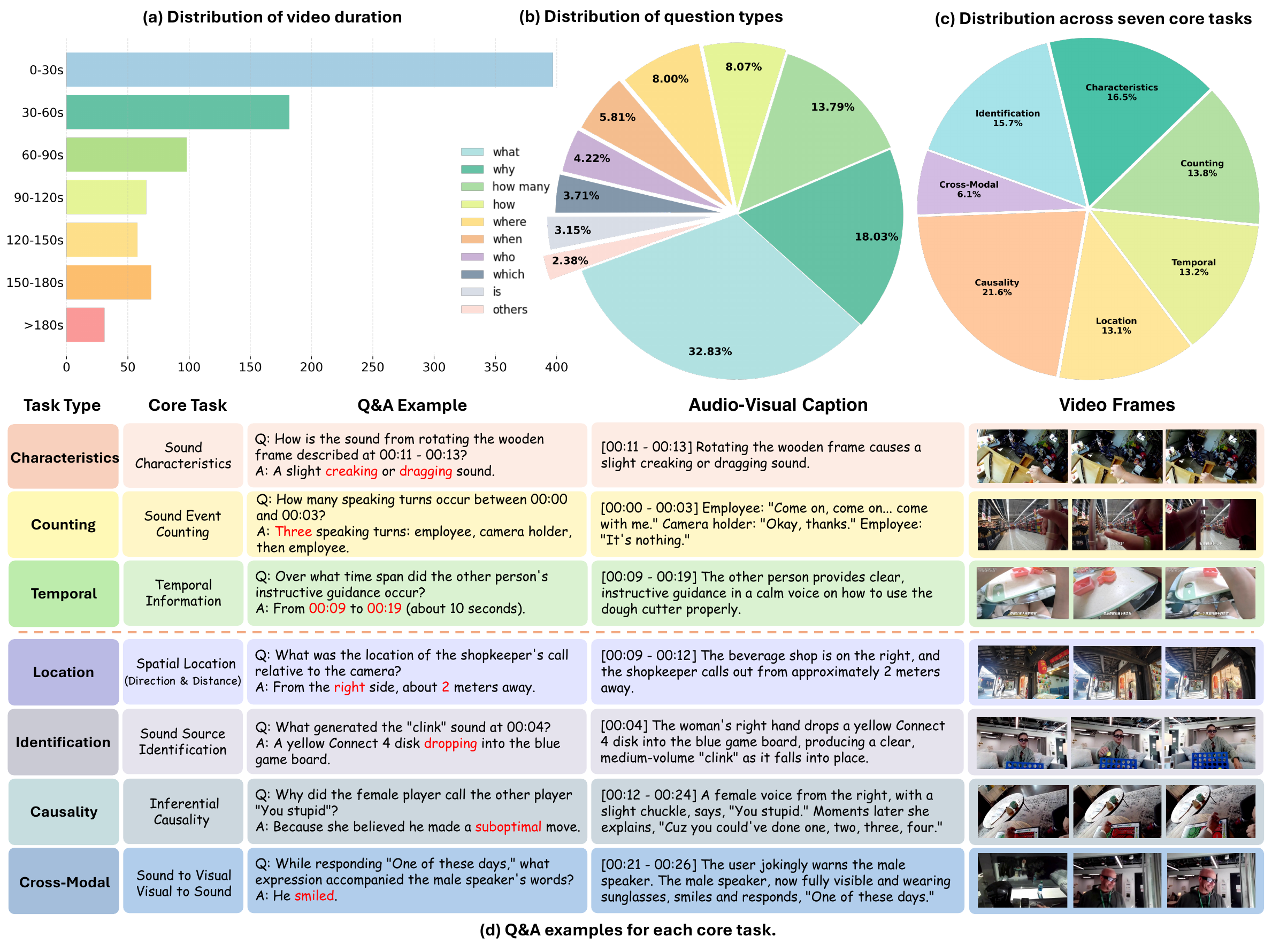}
  \caption{\textbf{Overview of the EgoSound task taxonomy and statistics.} (Top) Statistics on video length, question type, and the number of questions for each task category. (Bottom) A selection of representative examples for each core task of EgoSound.}
  \label{fig:overview}
\end{figure*}

\paragraph{Design Principles.} Our taxonomy follows three principles: a) Literature Grounding: We adapt established AudioQA and egocentric VideoQA tasks in Sec.~\ref{sec:related}, structuring them into sound-dependent vs. audio-visual categories; b) Comprehensive Assessment: We select diverse, complementary categories to ensure holistic evaluation; and c) Practical Relevance: We value tasks central to real-world scenarios aiming to support downstream applications. As illustrated in Fig.~\ref{fig:overview} (d), we curate seven egocentric sound tasks that target core capabilities essential for audio-visual understanding.
These tasks not only investigates the intrinsic properties of sounds, but also explores multimodal perception and reasoning. The intrinsic sound properties involve \textit{Sound Characteristics}, \textit{Counting}, and \textit{Temporal Attribute}, while the multimodal perception and reasoning aspects cover \textit{Spatial Location}, \textit{Sound Source Identification}, \textit{Inferential Causality}, and \textit{Cross-Modal Reasoning}. 

\begin{itemize}
\item \textbf{Sound Characteristics}. These tasks assess models' ability to describe the intrinsic acoustic properties of a sound, such as its perceived volume, texture, or timbre.
\item \textbf{Counting}. Counting tasks are designed to evaluate models' ability to track and enumerate distinct instances of auditory events or the repetitions of a specific sound. It also includes the number of times a particular word or phrase is mentioned in speech.
\item \textbf{Temporal Attribute}. Temporal tasks are proposed to evaluate models’ ability to analyze the temporal dynamics of a sound, including its duration, specific timing, and how its acoustic features evolve over time.
\item \textbf{Spatial Location}. Location tasks test models’ ability to localize a sound source in three-dimensional space relative to the egocentric observer, identifying both its direction and approximate distance.
\item \textbf{Sound Source Identification}. Identification tasks evaluate models' ability to identify the specific object or action that produced a sound, requiring the model to ground auditory signals to their corresponding visual events.
\item \textbf{Inferential Causality}. Causality tasks test models' higher-level ability to reason about the underlying cause or intent behind an auditory event by synthesizing information from the surrounding audio-visual context.
\item \textbf{Cross-Modal Reasoning}. Cross-Modal tasks assess a models' ability to integrate information across both modalities for complex inference. This includes using audio to interpret visual events (Audio-Guided Visual Reasoning) and using visual context to explain auditory events (Visual-Guided Audio Reasoning).
\end{itemize}

While the source egocentric video data provide meta annotation~\cite{grauman2022ego4d} or question-answer (Q\&A) pairs~\cite{xiao2025egoblind}, these annotations are predominantly focused on visual content. Consequently, they are insufficient for our primary goal of facilitating a deep and nuanced evaluation of sound event understanding. To address this gap, we designed a multi-stage data curation pipeline to generate high-quality, audio-centric Q\&A Pairs for EgoSound dataset. We illustrate our data curation pipeline in Fig.~\ref{fig:pipeline}. Our pipeline consists of three key stages: Human Interaction Annotation, Audio-Visual Caption Generation, and Q\&A Pairs Construction. 

\subsection{Data Curation Pipeline}
\paragraph{Human Interaction Annotation.}
Directly prompting an omni model for a general description of an egocentric video often results in captions that overlook significant scene details. Recognizing that physical interactions are the primary source for meaningful sound events, the first stage of our pipeline is designed to systematically annotate these key moments. To achieve this, we leverage the Qwen2.5-VL~\cite{bai2025qwen2} to perform automated annotation of human-object and human-human interactions. This process generates temporally-grounded labels that capture specific actions (e.g.
``At 3s, a girl in a red dress picks up the camera.",
%``At 15s, the user turns on the faucet.",
``From 45s to 48s, a man in a white shirt drives a black car past the camera wearer.").
These structured, interaction-focused annotations serve as a rich contextual foundation. They are used as conditioning prompts to guide the model in generating comprehensive audio-visual captions in the next stage, ensuring that the descriptions are anchored to the specific events that produce sound.

\paragraph{Audio-Visual Caption Generation.}
With egocentric sound understanding as the primary goal, the second stage of our pipeline generates audio–visual captions that center on sound while using visual human interaction labels as auxiliary context. Rather than describing the video broadly, we leverage the interaction annotations to disambiguate and refine the interpretation of audio, steering the model across seven sound-centric tasks (Sec.~\ref{subsec:task}). Concretely, for each annotated interaction, we prompt Gemini-2.5~\cite{comanici2025gemini} to describe the corresponding audio by linking each sound to its source, its acoustic traits, how many sources are active, when it occurs, how long it lasts, where it is in space, why it occurs, and how visual context helps explain audio. Additionally, all spoken words are also transcribed into the caption. Specific prompts are provided in the Supplementary. This yields captions grounded in first-person scenes but optimized for fine-grained, source-aware audio annotation, supporting high-quality audio–visual Q\&A in the final stage.

\paragraph{Q\&A Pairs Construction.}
To construct high-quality question–answer pairs while reducing potential hallucination effects in Gemini~\cite{comanici2025gemini}, we instruct GPT-4o~\cite{hurst2024gpt} to generate meaningful Q\&A samples based on detailed audio–visual captions and their corresponding video clip frames. We prompt GPT-4o~\cite{hurst2024gpt} to ask questions across seven core sound-centric tasks (Sec.~\ref{subsec:task}), with answers derived directly from the captions or inferred within their contextual bounds. As a double validation, every Q\&A pairs should be supported by visual evidence found in the video frames to ensure factual consistency.

To ensure the rigor of the evaluation and prevent guesswork, we adopted an open-ended question-answer (OpenQA) format, requiring the tested model to give descriptive answers rather than selecting from a list of options.
Following this pipeline, we constructed a total of 7315 validated QA pairs to comprehensively evaluate the audio-visual comprehension capabilities of MLLMs. 

\subsection{Dataset Statistics}
Tab.~\ref{tab:data} summarizes the data sources of EgoSound.
The dataset comprises 900 egocentric videos, including 640 clips from EgoBlind and 260 from Ego4D.
These videos vary in duration, ranging from 5 seconds to 5 minutes, with an average length of 59 seconds. The overall distribution of video durations is shown in Fig.~\ref{fig:overview} (a). In total, 7315 QA pairs are included, of which 4,969 are derived from EgoBlind, and the remaining ones from Ego4D. The proportions of QA pairs corresponding to the seven task categories are illustrated in Fig.~\ref{fig:overview} (c).
We further analyze the distribution of question types, 
%where ``what”, ``why”, and ``how many” emerge as the three most frequent categories, 
as shown in Fig.~\ref{fig:overview} (b).
\begin{table}[h]
\centering\small
  \begin{tabular*}{\columnwidth}{@{\extracolsep{\fill}}l|ccc}
    \toprule
    \textbf{Dataset} &  \textbf{Clips}  & \textbf{QA Pairs}  & \textbf{Dur.(s)} \\
    \midrule
    EgoBlind~\cite{xiao2025egoblind} & 640 & 4969 & $40.5$ \\
    Ego4D~\cite{grauman2022ego4d} & 260 & 2346 & $105.6$ \\
    \midrule
    \cellcolor{cyan!10}EgoSound  & \cellcolor{cyan!10}900 & \cellcolor{cyan!10}7315 & \cellcolor{cyan!10}$59.3$ \\
    \bottomrule
  \end{tabular*}
  %\vspace{-5pt}
\caption{Data source statistics for EgoSound.}
\label{tab:data}
\end{table}

\subsection{Human Verification}
\label{subsec:verification}
From over 7000 QA pairs, we balancedly sampled 350 QA pairs to conducted human verification study, with 50 pairs for each of the seven tasks. To ensure fairness and diversity, we prioritized videos containing a wide range of question types and maintained variation in average video duration during selection. Annotators verified the correctness of the QA with video alignment and assigned a quality score ($0 \text{–} 5$). 
Results confirm the reliability of our pipeline, achieving 92.1\% accuracy and an average score of 4.3. 

\section{Experiments}
\label{sec:experiments}
\begin{table*}[h]
\centering
\resizebox{\textwidth}{!}{%
\begin{tabular}{@{}l|ccc|cccc|r@{}}
\toprule
\textbf{Methods} & \textbf{Characteristics} & \textbf{Counting} & \textbf{Temporal} & \textbf{Location} & \textbf{Identification} & \textbf{Causality} & \textbf{Cross-Modal} & \textbf{Average}\\
\midrule
\rowcolor{gray!20} % Gray background for the Human row
\textbf{Human} & 95.6 / 4.4 & 82.5 / 4.0 & 67.5 / 3.3 & 81.6 / 3.8 & 69.7 / 3.6 & 95.2 /4.3 & 90.5 / 4.1 & 83.9 / 3.9 \\
\midrule
\multicolumn{9}{@{}c}{\textit{Open-source Models}} \\
\midrule
VideoLLaMA2.1-AV-7B~\cite{cheng2024videollama} & 15.5 / 1.3 & 27.7 / 1.4 & 29.1 / 1.7 & 21.7 / 1.2 & 19.4 / 1.1 & 16.5 / 1.1 & 13.6 / 0.9 & 20.5 / 1.3\\
video-SALMONN 2+ -7B ~\cite{tang2025video} & 29.2 / 2.0 & 40.1 / 2.1 & 31.5 / 1.8 & 26.9 / 1.5 & 34.1 / 1.8 & 46.9 / 2.5 & 40.6 / 2.2 & 36.0 / 2.0 \\
video-SALMONN 2+ -72B~\cite{tang2025video} & 38.4 / 2.5 & 51.3 / 2.8 & 37.3 / 2.1 & 35.4 / 1.9 & 43.7 / 2.3 & 63.5 / 3.3 & 51.1 / 2.8 & 46.6 / 2.5 \\
Qwen2.5-Omni-3B ~\cite{xu2025qwen2} & 28.3 / 1.9 & 48.7 / 2.5 & 31.5 / 1.8 & 26.7 / 1.5 & 17.7 / 1.0 & 46.8 / 2.4 & 33.7 / 1.9 & 33.9 / 1.9 \\
Qwen2.5-Omni-7B~\cite{xu2025qwen2} & 34.1 / 2.1 & \textbf{56.3} / 2.9 & 31.4 / 1.7 & 30.7 / 1.6 & 23.4 / 1.3 & 55.3 / 2.8 & 42.6 / 2.3 & 39.8 / 2.1 \\ 
MiniCPM-o 2.6-8B~\cite{yao2024minicpm} & 33.5 / 2.2 & 50.7 / 2.7 & 36.4 / 2.0 & 26.5 / 1.4 & 41.6 / 2.2 & 48.6 / 2.5 & 42.2 / 2.3 & 40.4 / 2.2 \\
Qwen3-Omni-Instruct-30B~\cite{xu2025qwen3} & \underline{45.8} / 2.8 & \underline{55.0} / 2.9 & \textbf{50.8} / 3.0 & \underline{40.1} / 2.2 & \underline{49.8} / 2.6 & \underline{65.2} / 3.4 & \underline{51.6} / 2.8 & \underline{51.9} / 2.8 \\
Qwen3-Omni-Thinking-30B~\cite{xu2025qwen3} & \textbf{55.0} / 3.1 & 49.0 / 2.6 & \underline{46.7} / 2.7 & \textbf{50.0}/ 2.6 & \textbf{59.5} /3.0 & \textbf{73.3} / 3.7 & \textbf{54.5} / 2.9 & \textbf{56.7} / 3.0 \\
\midrule
\multicolumn{9}{@{}c}{\textit{Egocentric Models}} \\
\midrule
EgoGPT-7B~\cite{yang2025egolife} & 21.2 / 1.8 & 52.6 / 2.8 & 38.7 / 2.2 & 26.1 / 1.5 & 26.7 / 1.4 & 41.9 / 2.2 & 28.9 / 1.7 & 34.3 / 2.0 \\
\bottomrule
\end{tabular}}
\caption{\textbf{Evaluation results of MLLMs on EgoSound.} The best results are marked in \textbf{bold}, and the second-best are \underline{underlined}. Characteristics, Counting, and Temporal measure the model's ability to perceive the intrinsic properties of sound. Location, Causality, Reasoning, and Cross-Modal go beyond this, further evaluating the model's multimodal perception and reasoning capabilities.}
\label{tab:result}
\end{table*}

In this section, we present our experimental results. We describe our experimental setup in Sec.~\ref{sub:setup}, where we introduce the models used and the evaluation setting. The main results are presented in Sec.~\ref{sub:result}. 
Finally, we further discuss the audio-only evaluation in Sec.~\ref{sub:audio-only}.

\subsection{Experimental Setup}
\label{sub:setup}
\paragraph{Evaluated MLLMs. }
We evaluate our benchmark on a range of state-of-the-art omni-models that can jointly process audio and video signals. Specifically, the evaluated models include VideoLLaMA2.1-AV~\cite{cheng2024videollama}, video-SALMONN 2+ (7B, 72B)~\cite{tang2025video}, MiniCPM-o 2.6~
\cite{yao2024minicpm}, Qwen2.5-Omni (3B, 7B)~\cite{xu2025qwen2}, Qwen3-Omni-30B (Instruct, Thinking)~\cite{xu2025qwen3}.
These models range in size from 3B to 72B parameters, covering a broad spectrum of model capacities.
In addition, we evaluate EgoGPT~\cite{yang2025egolife}, a model specifically tailored for egocentric video understanding.
For fairness, we exclude Gemini~\cite{comanici2025gemini} from evaluation, since the captions in our dataset were annotated using Gemini 2.5 Flash, which may introduce potential bias.

\paragraph{Evaluation Metrics.}
Since the QA pairs are in open-ended form, we employ GPT-5~\cite{openai_gpt5_system_card_2025} as an automatic validation tool to assess model performance on EgoSound.
GPT-5~\cite{openai_gpt5_system_card_2025} evaluates the factual consistency between each model’s predicted answer and the ground-truth reference (correct answer).
Following prior work~\cite{cheng2024videollama,xiao2025egoblind}, we define two evaluation metrics for model predictions: Accuracy ($0 \text{–} 100\%$), the percentage of predicted answers judged as ``correct"; and Score ($0 \text{–} 5$), the degree of semantic consistency between the predicted and reference answers, where 5 indicates a fully correct answer and 0 is completely wrong. 

\paragraph{Human Evaluation.}
We recruited two English-proficient evaluators, each holding a bachelor’s degree and possessing solid research experience in computer vision, to conduct the human evaluation of our benchmark. Following the protocol in Sec.~\ref{subsec:verification}, we sampled a subset of 350 QA pairs for evaluation. Additionally, we randomly shuffled the QA pairs to prevent similar questions from appearing consecutively. The human evaluation results are shown in Tab.~\ref{tab:result}.
%thereby reducing potential evaluation bias.

\subsection{Main Results}
\label{sub:result}
We benchmark nine representative MLLMs on the EgoSound, and the quantitative results are summarized in Tab.~\ref{tab:result}. Overall, our experiments reveal that egocentric sound understanding remains a formidable challenge for current MLLMs, despite their strong progress in vision–language integration.

\textbf{\ding{172} EgoSound poses a significant challenge in current MLLMs.}
Human evaluators achieve an average accuracy of 83.9\%, whereas the best model, Qwen3-Omni-Thinking-30B~\cite{xu2025qwen3}, reaches only 56.7\%, indicating a large gap of over 27 points. This demonstrates that while MLLMs can align vision and language, they still struggle to ground sound cues for reliable perception and reasoning. 
The performance gap confirms the unique difficulty of multisensory understanding in first-person settings, where sound and vision are deeply entangled.

\textbf{\ding{173} Well-validated ability drops on audio.} 
Existing MLLMs are well-known for their strong perception ability within the visual domain, such as recognizing object characteristics or identifying approximate locations. However, when evaluated on audio-centric tasks, these well-validated abilities show a clear performance degradation. For instance, models including VideoLLaMA2.1-AV-7B~\cite{cheng2024videollama}, Video-SALMONN 2+~\cite{tang2025video}, Qwen2.5-Omni~\cite{xu2025qwen2}, and MiniCPM-o~\cite{yao2024minicpm} perform notably worse on tasks involving sound characteristics and spatial localization, often even below their results on counting, causality, or cross-modal reasoning. 
This reveals that despite supporting audio inputs, current MLLMs still lag significantly in fine-grained auditory perception.

\textbf{\ding{174} Model scale's impact on performance.}
Larger models generally show improved performance, but this scaling advantage does not eliminate all performance gaps. For instance, video-SALMONN 2+ 72B~\cite{tang2025video} significantly outperforms its 7B counterpart (46.6\% vs. 36.0\%), and Qwen2.5-Omni-7B~\cite{xu2025qwen2} exceeds the 3B version by 5.9\% percentage points. Similarly, models with a larger scale (e.g., 30B) perform better than those in the smaller scale range (3B/7B/8B). This is likely because larger models are often trained with more diverse data, leading to better generalization ability. However, this trend is not absolute. The performance difference between VideoLLaMA2.1-AV-7B~\cite{cheng2024videollama} and Qwen2.5-Omni-3B~\cite{xu2025qwen2}, as well as the fact that Video-SALMONN 2+ 72B~\cite{tang2025video} fails to surpass Qwen3-Omni-Instruct/Thinking-30B~\cite{xu2025qwen3}, clearly indicates that simply increasing model size does not guarantee superior results.

\textbf{\ding{175} Egocentric pretraining does not help.}
The EgoGPT-7B model~\cite{yang2025egolife}, which is specifically adapted for egocentric data, achieves an average accuracy of 34.3\%, substantially lower than the best Qwen3-Omni-Instruct/Thinking-30B~\cite{xu2025qwen3} models (above 50\%) and even inferior to other models of comparable scale. Although this is not an absolutely fair comparison, the relatively poor performance of EgoGPT-7B suggests that pretraining or finetuning on egocentric data alone does not directly address models’ limitations in sound-related understanding. This observation further underscores the importance of developing modality-balanced MLLMs that can jointly reason over vision, language, and audio. We also note that a broader conclusion would ideally rely on multiple egocentric models; however, the current availability of open-source egocentric models remains limited. This again highlights the need for more egocentric-oriented MLLMs to advance research in multisensory egocentric understanding.

\subsection{Audio-Only Evaluation}
\label{sub:audio-only}
\begin{figure}[!t]
    \centering
    \includegraphics[width=\linewidth]{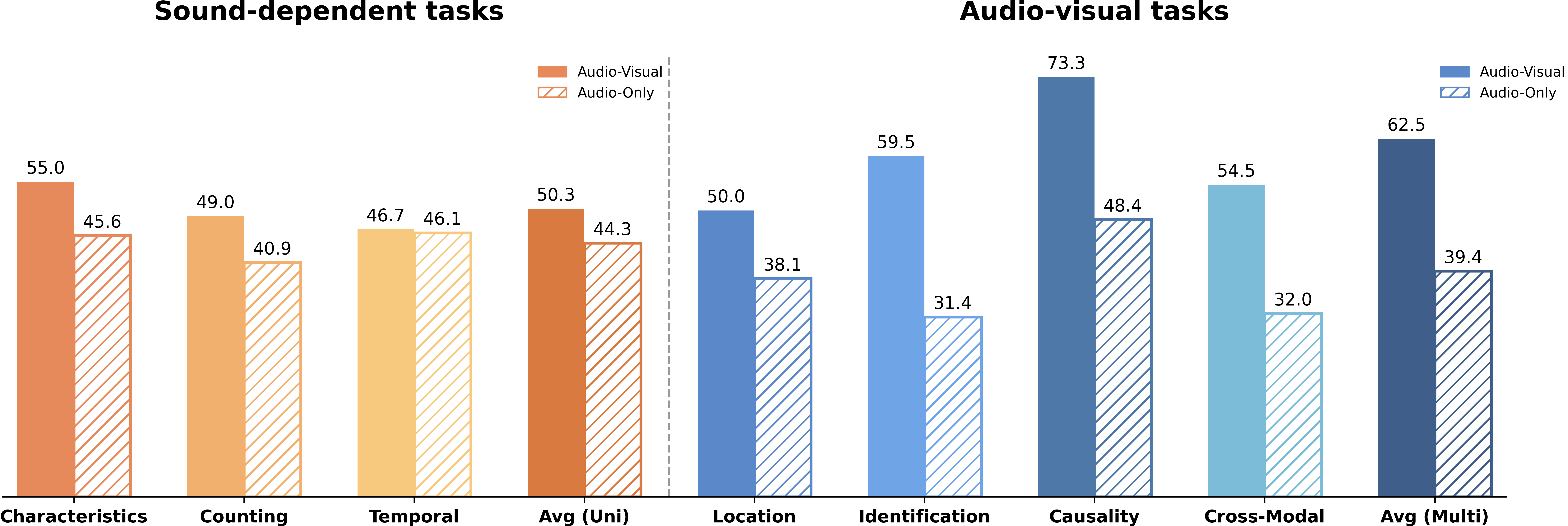}
    \caption{\textbf{Accuracy comparison on EgoSound for Qwen3-Omni-Thinking~\cite{xu2025qwen3} with audio–visual \textit{vs} audio-only input.} Sound-dependent three tasks results are shown in the left (\textcolor{orange}{orange}), while the right (\textcolor{cyan}{blue}) shows the results for other four tasks that depend on both visual and audio input.}
    \label{fig:audio}
    \vspace{-0.05in}
\end{figure}

To investigate the role of visual information in EgoSound, we conduct an audio-only ablation that removes visual inputs while keeping all other settings identical. Specifically, we feed the same questions to the same model, but instead of providing both RGB frames and audio, we retain only the audio track during inference. Following the results in Tab.~\ref{tab:result}, we select the best-performing model, i.e., Qwen3-Omni-Thinking-30B~\cite{xu2025qwen3}, to perform this ablation. A comparison between dual-modality (audio + visual) and single-modality (audio-only) performance is summarized in Fig.~\ref{fig:audio}.

\begin{figure*}[!t]
    \centering
    \includegraphics[width=\linewidth]{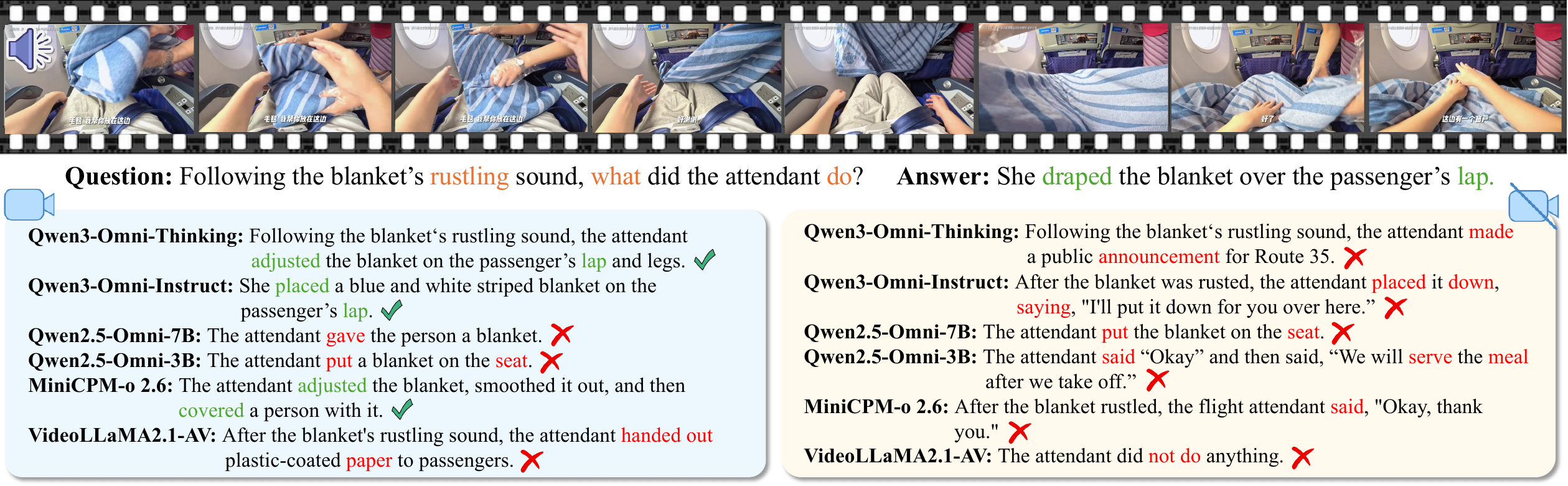}
    %\vspace{-5pt}
    \caption{\textbf{Comparison of Cross-Modal Reasoning with and without visual input.} The video shows an egocentric airplane scene in which a flight attendant handles a blanket for the passenger. The question asks what happens after the rustling sound produced during this action. The left side presents model outputs with audio-visual frames; the right side presents outputs with audio alone.}
    \label{fig:comparison}
\end{figure*}
From the results, we highlight two main observations: 1) \textbf{Sound-dependent tasks remain stable.} For sound-only QA tasks
(Characteristics, Counting, and Temporal)
, the model maintains relatively stable performance, with only slight drops from 50.3\% to 44.3\% on average. This is consistent with intuition: when a task depends solely on auditory cues, removing visual inputs should not substantially affect model accuracy. 2) \textbf{Audio-visual tasks degrade significantly.} For tasks requiring joint reasoning 
(Location, Identification, Causality, and Cross-Modal)
, we observe a significant decline in performance under the audio-only condition. For example, performance dropped by 28.1 and 24.9 points in Localization and Causality respectively, and the average accuracy also decreased by more than 20\%. These findings validate that our audio-visual tasks indeed require integrated reasoning over both modalities, rather than relying on audio alone. Finally, we note that although the model’s performance on audio-visual tasks drops noticeably under audio-only input, it does not completely fail. This mirrors the human condition where visually impaired individuals can still interpret the world through auditory cues alone, suggesting that partial reasoning remains possible even without visual modality.

\noindent\textbf{Further analysis of the visual modality.}
%As shown in Fig.~\ref{fig:audio}, 
% even the best model degrades notably on visual-audio tasks when visual inputs are removed. 
To further illustrate why visual signals is essential for joint reasoning, we visualize representative examples in Fig.~\ref{fig:comparison}, comparing model predictions under different input modalities.
With dual-modality input (left), the strongest models (Qwen3-Omni-Thinking~\cite{xu2025qwen3}, Qwen3-Omni-Instruct~\cite{xu2025qwen3}, and MiniCPM-o 2.6~\cite{yao2024minicpm}) successfully associate the rustling sound with the moment the blanket is being adjusted and correctly ground the subsequent action, identifying that the attendant places or drapes the blanket over the passenger’s lap. In contrast, under the audio-only setting (right), all of these models fail to provide the correct answer. This gap clearly demonstrates the importance of cross-modal cooperation for comprehensive first-person scene understanding.
%highlighting the value of our proposed EgoSound benchmark. 
Additionally, we observe that weaker models fail under both input conditions, often mislocalizing the action or hallucinating unrelated events. These observations confirm that our QA tasks pose significant challenges to current MLLMs and serve as a realistic and demanding testbed for multisensory egocentric reasoning.

\section{Conclusion}
\label{sec:conclusion}
In this work, we introduced the EgoSound, the first benchmark designed to systematically evaluate egocentric sound understanding in Multimodal Large Language Models. By unifying the data from Ego4D and EgoBlind and establishing a seven-task taxonomy,
spanning sound characteristics, sound event counting, temporal information, spatial location, sound source identification, inferential causality, and cross-modal reasoning, 
EgoSound offers a comprehensive and realistic testbed for multisensory egocentric intelligence. Through large-scale evaluation across nine state-of-the-art MLLMs, we reveal that while existing models exhibit emergent auditory reasoning abilities, they continue to struggle with fine-grained audio perception and first-person multimodal joint reasoning. Our audio-only ablation studies further highlight the indispensable role of visual cues in many sound-related reasoning scenarios, reinforcing the need for balanced multimodal learning across vision, audio, and language. EgoSound aims to spark deeper exploration into multisensory modeling, especially in the first-person perception domain, for boosting better and robust human-aligned intelligence. 
We hope this benchmark not only exposes current limitations but also catalyzes progress toward next-generation MLLMs that can truly see, hear, and understand the world from a first-person view.

\section*{Acknowledgments}
This work is supported by TeleAI.

{   %\clearpage 
    \small
    \bibliographystyle{ieeenat_fullname}
    \bibliography{main}
}

% WARNING: do not forget to delete the supplementary pages from your submission 
\clearpage
\setcounter{page}{1}
\maketitlesupplementary
\appendix
\section{More Construction Details}
\label{sec:detail}
This section provides additional details on the prompting strategies used in our multi-stage data curation pipeline. As described in Sec.3.3 of the main paper, EgoSound relies on structured human–interaction annotations, fine-grained audio–visual caption generation, and a robust QA construction process. Below, we present the exact prompt designs that operationalize these stages.
\begin{figure}[h]
    \centering
    \includegraphics[width=\linewidth]{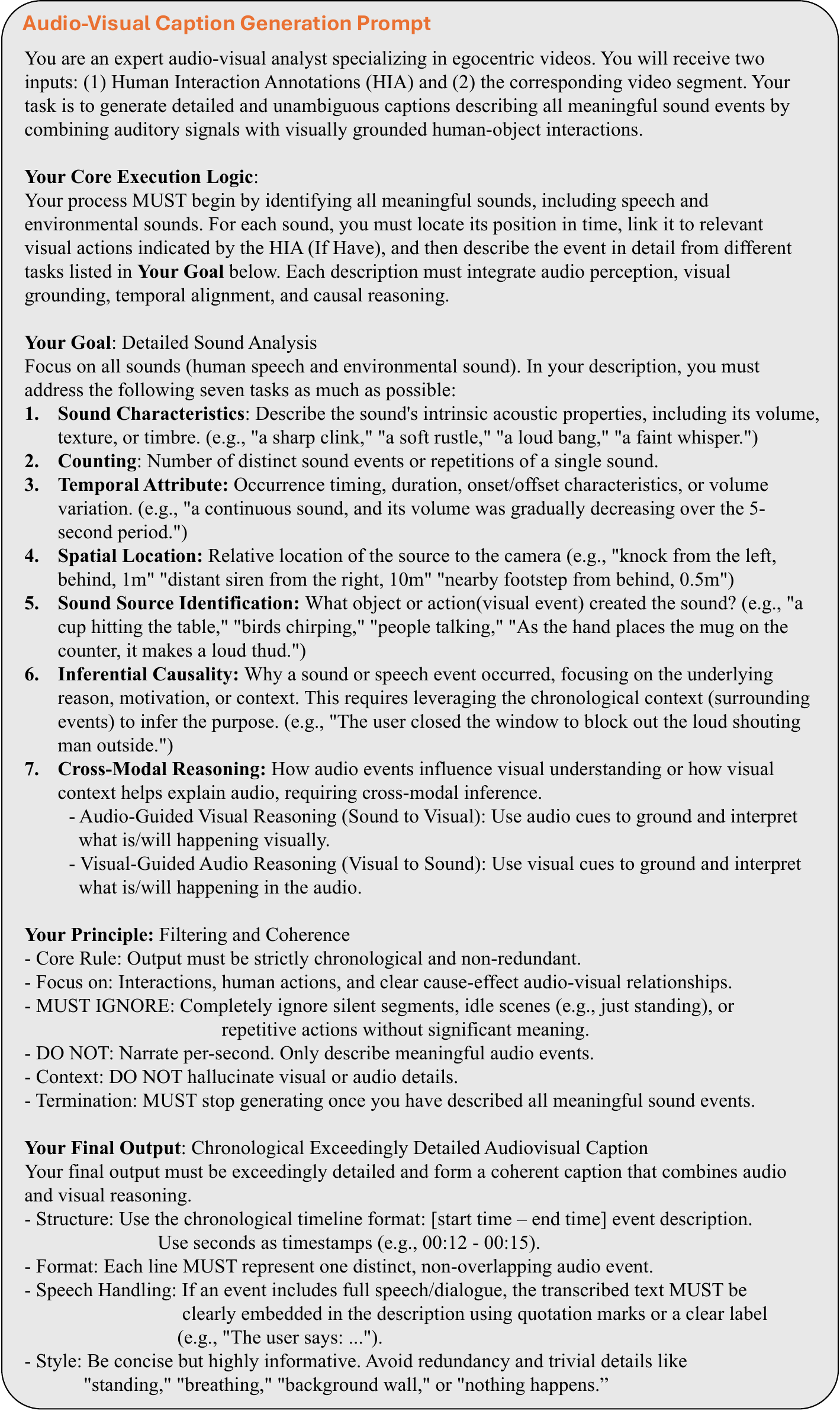}
    \caption{\textbf{Prompt of Audio-Visual Caption Generation.}}
    \label{fig:caption}
\end{figure}

\subsection{Prompt for Caption Generation}
To transform annotated human–object and human–human interactions into detailed, sound-centric descriptions, we follow~\cite{ma2025omni} and design a specialized prompt that instructs the model to generate chronological audio–visual captions grounded in both audio cues and visual context. Fig.~\ref{fig:caption} shows the full prompt template used for audio–visual caption generation.

\subsection{Prompt for QA Pairs Generation}
\begin{figure}[!t]
    \centering
    \includegraphics[width=\linewidth]{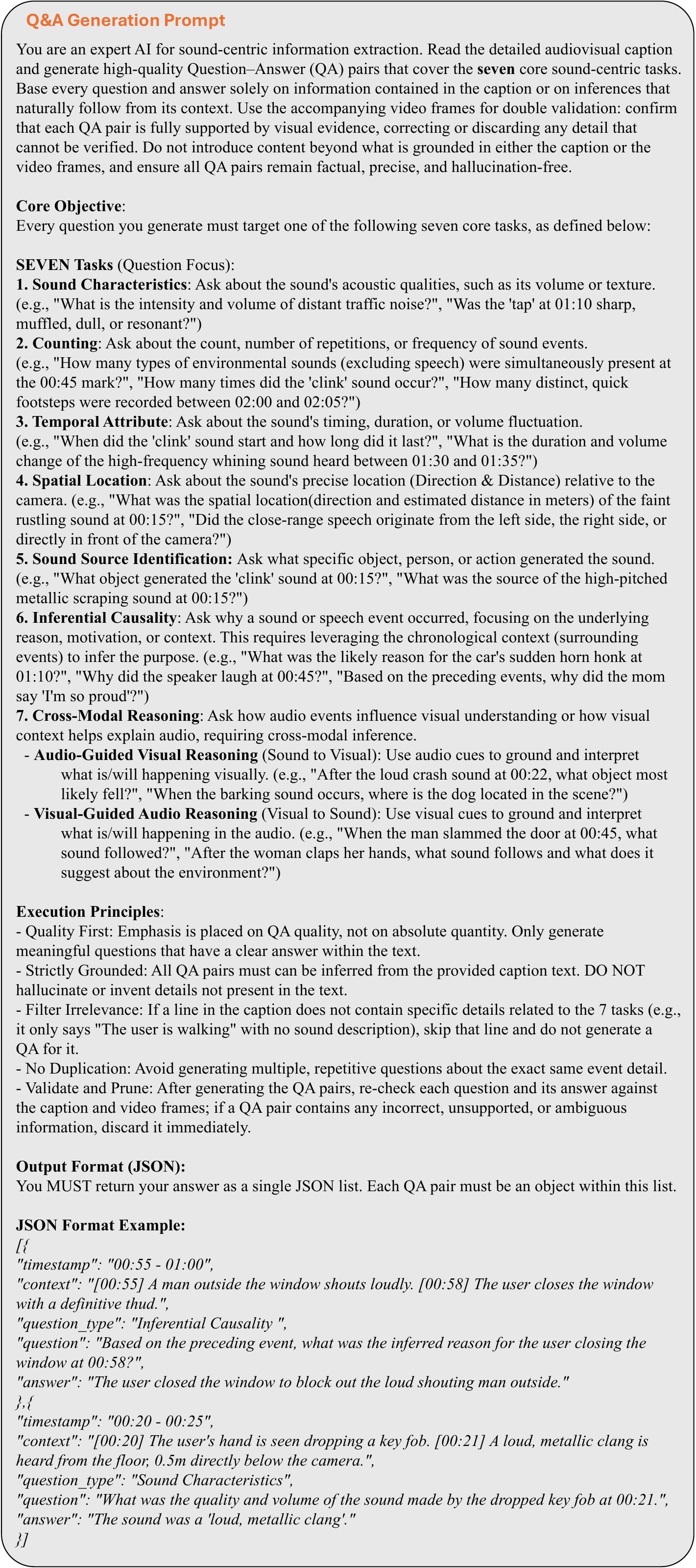}
    \caption{\textbf{Prompt of QA Pairs Generation.}}
    \label{fig:qa}
    % \vspace{-0.05in}
\end{figure}

Building on the generated captions, we design a new prompt to construct sound-centric QA pairs. The prompt instructs the model to produce questions spanning the seven core tasks defined in EgoSound, ensuring that the resulting QA pairs cover intrinsic audio perception, spatial reasoning, causal inference, and cross-modal understanding. Fig.~\ref{fig:qa} presents the complete prompt used for QA construction.

\section{More Analysis and Visualization Results}
\label{sec:analysis}
In this section, we provide additional analyses and visualizations to complement the results reported in the main paper. Beyond quantitative comparisons, these materials offer deeper insight into how current MLLMs behave across different sound-centric tasks in EgoSound. We include the implementation details of our main experiments, the full prompt used for OpenQA evaluation, and representative QA examples from both Ego4D and EgoBlind to illustrate the diversity and complexity of our data sources. We also present a range of failure cases that reveal systematic weaknesses in temporal grounding, spatial localization, sound source identification, cross-modal alignment, and acoustic perception. Together, these analyses provide a more comprehensive understanding of the challenges posed by egocentric sound reasoning and highlight key opportunities for improving future multimodal models.

\subsection{Implement Details of Experiments}
For fair and reproducible evaluation, we execute different model sizes on hardware appropriate to their computational requirements. Specifically, all small and medium-sized models—including VideoLLaMA2.1-AV-7B~\cite{cheng2024videollama}, video-SALMONN 2+-7B~\cite{tang2025video}, Qwen2.5-Omni-3B~\cite{xu2025qwen2}, Qwen2.5-Omni-7B~
\cite{xu2025qwen2}, and MiniCPM-o 2.6-8B~\cite{yao2024minicpm}—are run on NVIDIA RTX 4090 GPUs. Larger models—including video-SALMONN 2+-72B~\cite{tang2025video}, Qwen3-Omni-Instruct-30B~\cite{xu2025qwen3}, and Qwen3-Omni-Thinking-30B~\cite{xu2025qwen3}—are executed on NVIDIA H200 GPUs to ensure stable inference and prevent memory bottlenecks. All models are evaluated under the same zero-shot, single-round inference protocol. For visual inputs, video frames are sampled at dataset-specific rates between 0.5–1 fps, and no maximum frame limit is enforced, preserving the temporal continuity of each egocentric sequence. Audio streams are fed at their original sampling rate and are not modified. Each model receives identical multimodal inputs consisting of the sampled video frames, synchronized audio, and the question text.

\subsection{Prompt for OpenQA Evaluation}
\begin{figure}[!t]
    \centering
    \includegraphics[width=\linewidth]{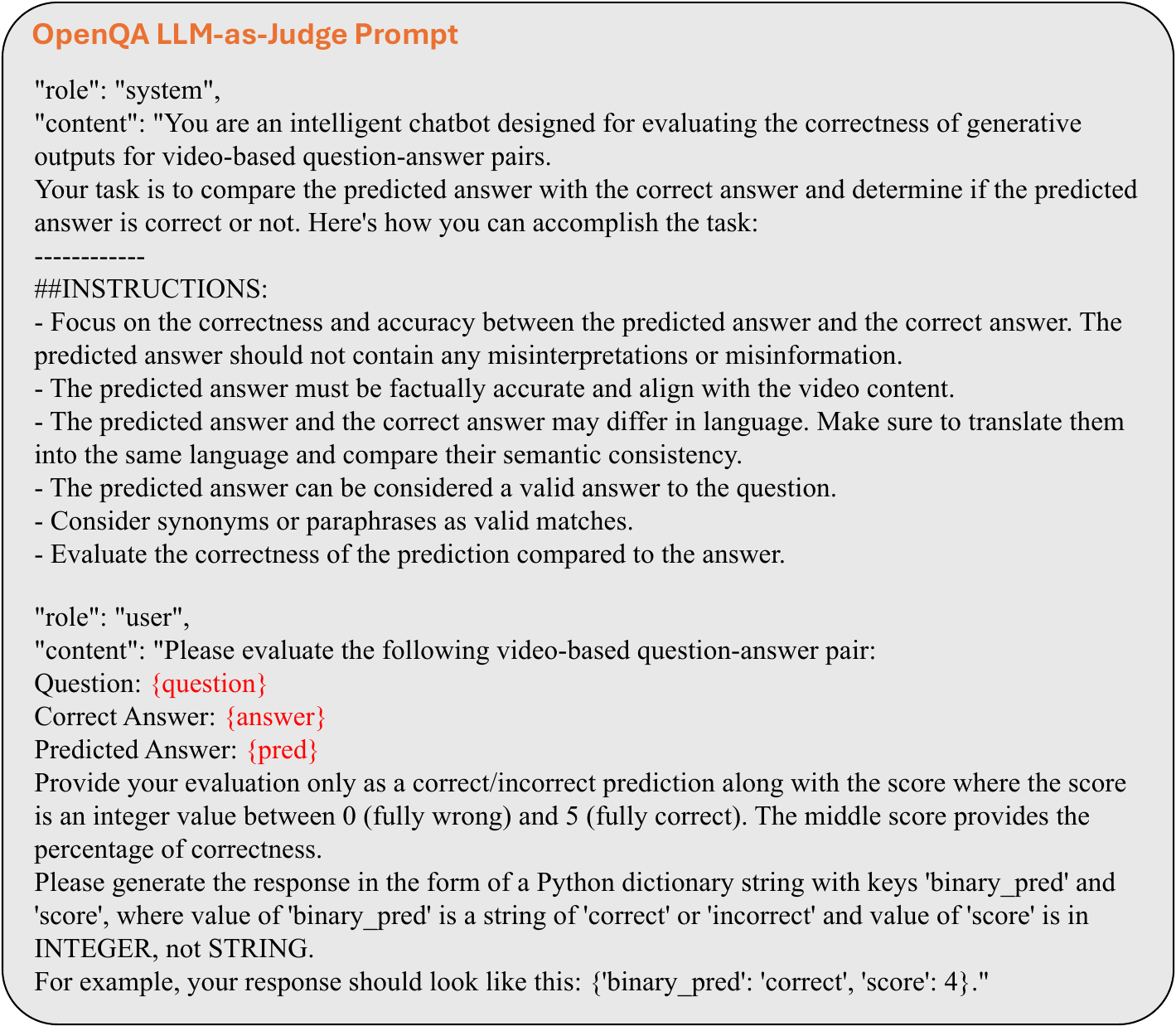}
    \caption{\textbf{Prompt of LLM-as-Judge.} The prompt takes as input the \textit{question}, the correct answer \textit{(answer)}, the model’s prediction \textit{(pred)}, to produce the resulting evaluation.}
    \label{fig:judge}
\end{figure}

Given the subjective nature of open-ended responses, we adopt GPT-5~\cite{openai_gpt5_system_card_2025} as an automated judge to provide consistent and scalable evaluation. The LLM judge assesses the factual consistency of each model prediction relative to the ground truth, following the prompt design illustrated in Fig.~\ref{fig:judge}.

\subsection{QA examples Visualization}
\begin{figure*}[!t]
    \centering
    \includegraphics[width=\linewidth]{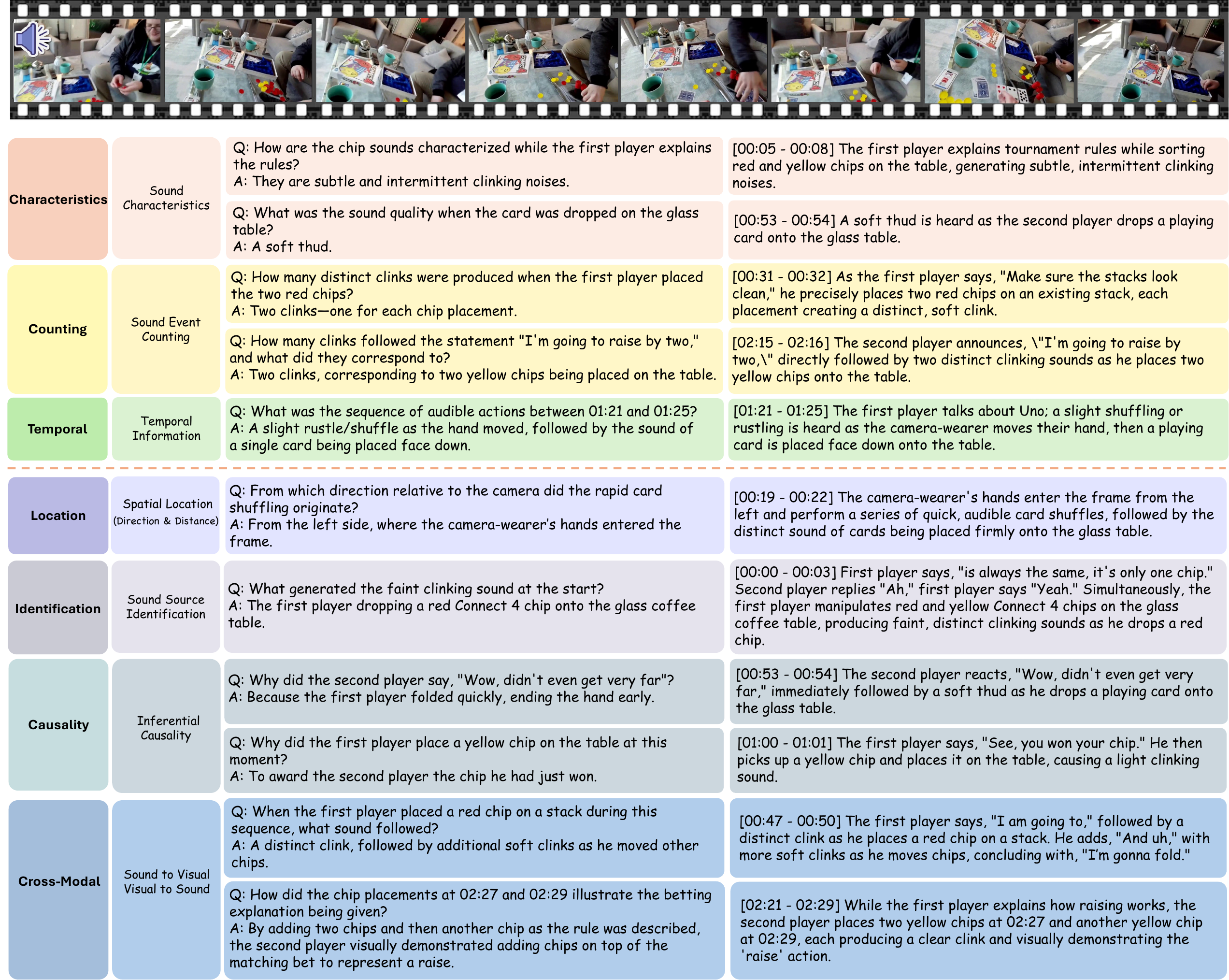}
    \caption{A visualization of representative QA examples, the video source is from the EGO4D~\cite{grauman2022ego4d} dataset.}
    \label{fig:vis1}
\end{figure*}

\begin{figure*}[!t]
    \centering
    \includegraphics[width=\linewidth]{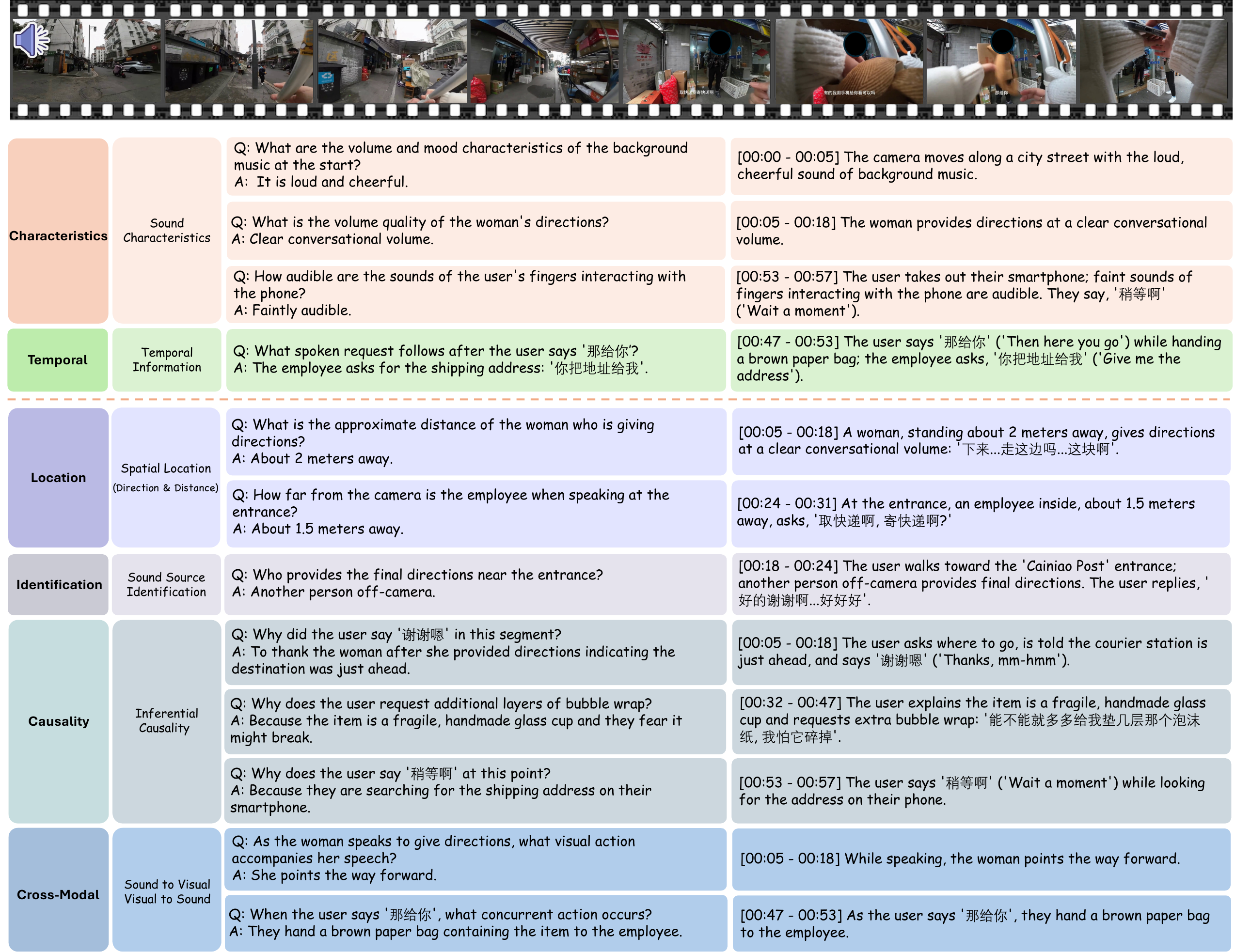}
    \caption{A visualization of representative QA examples, the video source is from the EgoBlind~\cite{xiao2025egoblind} dataset.}
    \label{fig:vis2}
\end{figure*}

Fig.~\ref{fig:vis1} and Fig.~\ref{fig:vis2} present representative QA examples from the Ego4D~\cite{grauman2022ego4d} and EgoBlind~\cite{xiao2025egoblind} subsets of EgoSound. Each QA pair is constructed from both the detailed clip caption shown in the rightmost column and the corresponding video frames, ensuring that every question and answer is grounded in the underlying audiovisual evidence. The examples follow the seven egocentric sound tasks defined in the main paper: the intrinsic sound properties (\textit{Sound Characteristics}, \textit{Counting}, \textit{Temporal Attribute}) and the multimodal perception and reasoning tasks (\textit{Spatial Location}, \textit{Sound Source Identification}, \textit{Inferential Causality}, \textit{Cross-Modal Reasoning}). These visualizations illustrate how the caption-to-QA pipeline faithfully anchors sound events to their temporally aligned egocentric audio–visual context.

\subsection{Failure Cases Analysis}
\begin{figure}[!t]
    \centering
    \includegraphics[width=\linewidth]{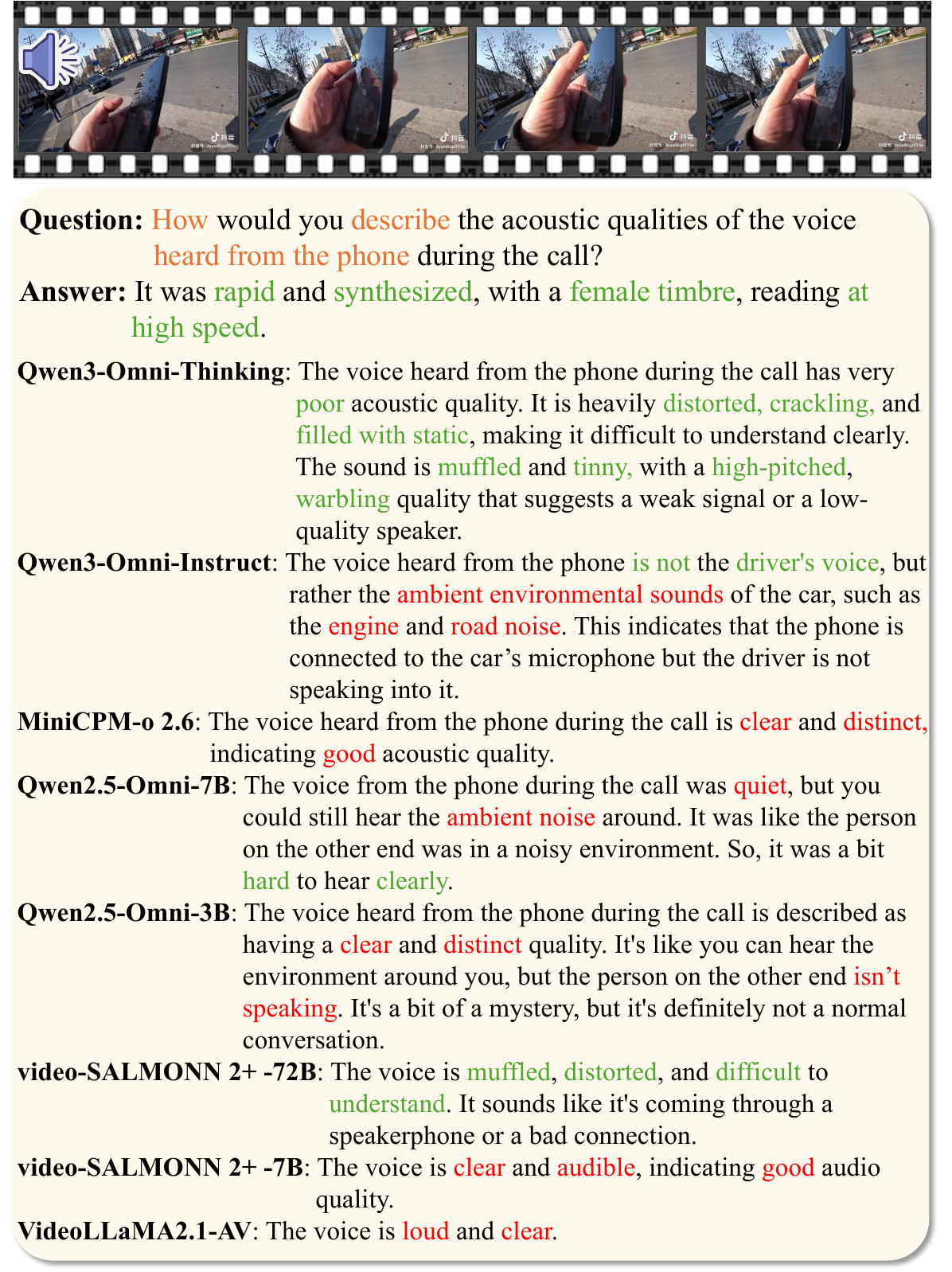}
    \caption{MLLMs fail to consistently perceive low-quality and distorted sound characteristics.}
    \label{fig:characteristic}
\end{figure}

To further understand the limitations of current MLLMs on EgoSound, we provide a detailed analysis of representative failure cases across the seven sound-centric tasks. The following examples highlight common error patterns—such as misaligned temporal grounding, incorrect spatial reasoning, audio–visual mismatches, off-screen sound misidentification, and difficulties in perceiving low-quality or rapid speech. 
\paragraph{Failure case of Sound Characteristics.}
Fig.~\ref{fig:characteristic} highlights a failure case in sound characteristics perception, where the voice from the phone is rapid, synthesized, and unclear. The models exhibit mixed results in handling this distorted audio. Some models detect the poor acoustic quality, describing the voice as distorted, muffled, and difficult to understand due to static or ambient noise, suggesting issues with the signal or the speaker. Others, however, interpret the voice as clear or loud, indicating good acoustic quality, despite the presence of distortions and background noise.

This case underscores the challenges faced by current MLLMs in accurately perceiving audio with poor quality. In everyday settings, audio quality is often inconsistent, with varying degrees of distortion and noise. The variety of responses shown in the outputs reveals that current MLLMs struggle to consistently interpret such challenging audio conditions. Therefore, EgoSound, which includes a range of real-life scenarios with varying audio quality, presents a significant challenge to the audio perception capabilities of current MLLMs.

\begin{figure}[!t]
    \centering
    \includegraphics[width=\linewidth]{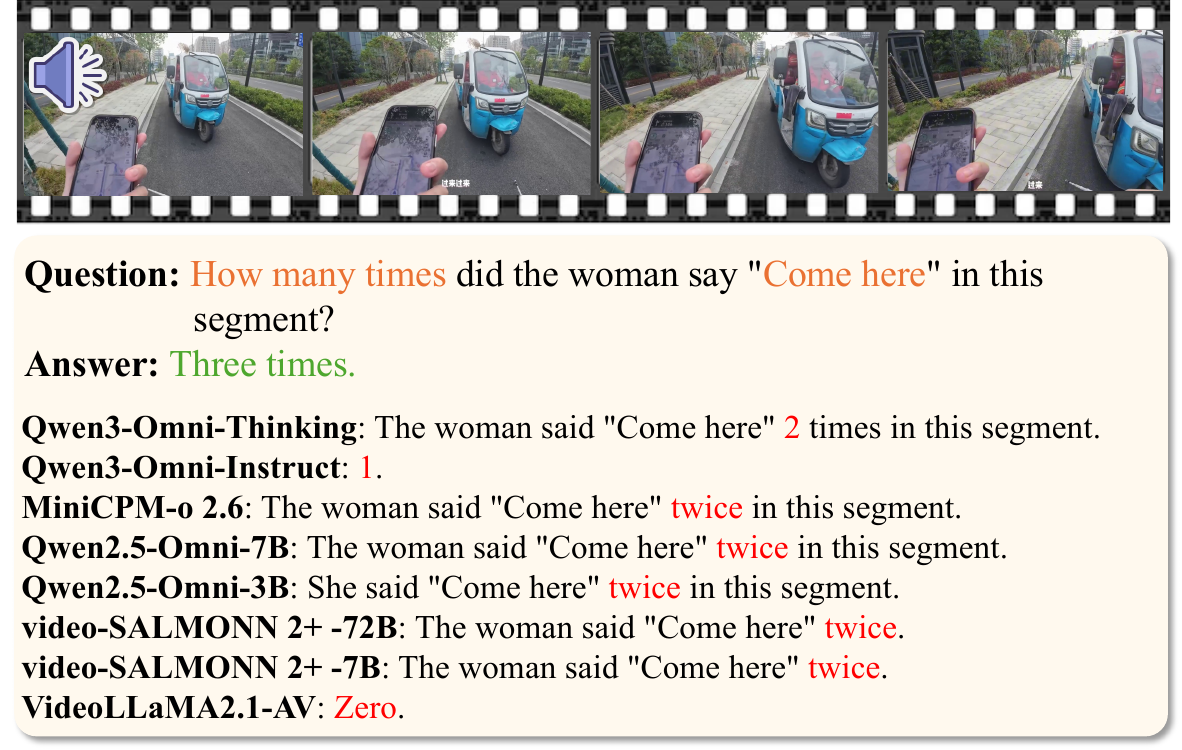}
    \caption{MLLMs fail to enumerate the repetitions of a specific phrase in fast-spoken or audio-unclear egocentric videos.}
    \label{fig:counting}
\end{figure}

\paragraph{Failure case of Counting.}
Fig.~\ref{fig:counting} illustrates a case where models struggle with detecting the counting of the phrase ``Come here" in a egocentric video segment. The correct answer is that the woman says ``Come here" three times in total, but most models fail to capture this accurately. While some models correctly identify two occurrences, others report only one or even zero. The main reason for this discrepancy is that the first instance of ``Come here" in the second frame is spoken too quickly and lightly for the models to detect, leading to its omission in their results.

This example highlights the challenges faced by current MLLMs in detecting fast-spoken or unclear audio, even in relatively simple scenarios. The inability to capture rapid speech accurately, particularly in noisy or dynamic environments, poses significant obstacles for the audio perception capabilities of these models. 

\begin{figure}[!t]
    \centering
    \includegraphics[width=\linewidth]{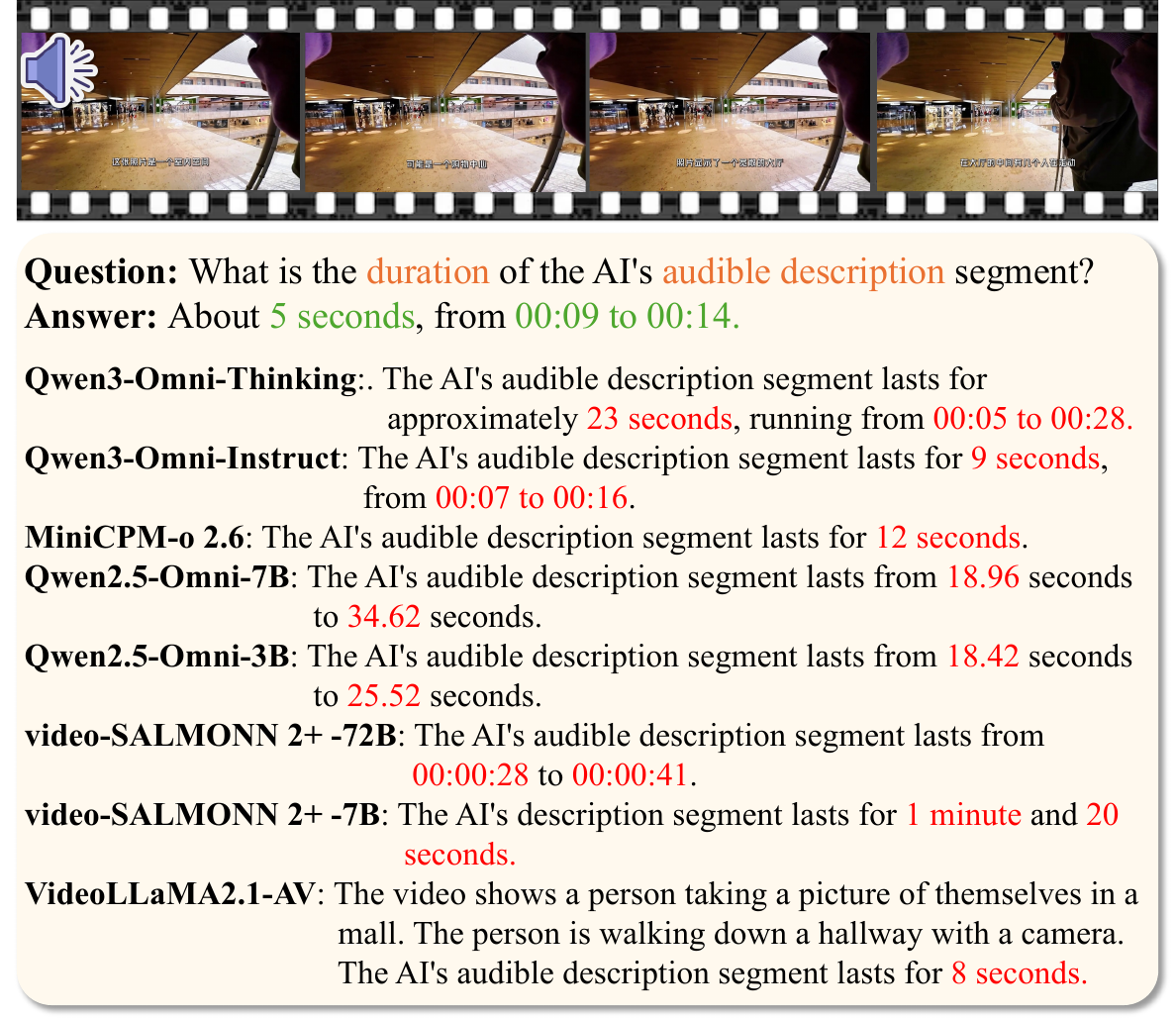}
    \caption{MLLMs fail to accurately identify temporal boundaries of sound events in short egocentric videos.}
    \label{fig:temporal}
\end{figure}

\paragraph{Failure case of Temporal Attribute.}
Fig.~\ref{fig:temporal} illustrates a failure case in the temporal attribute task, where models struggle to accurately identify the duration of the sound event. The correct duration of the audible description is from 00:09 to 00:14, lasting approximately 5 seconds. 
However, models fail to align the audible description temporally, often outputting timestamps that extend beyond the actual video length.
For example, some models incorrectly predict durations as long as 23 seconds or more, even though the entire video is only 16 seconds long. This error indicates that the MLLMs fail to accurately align the auditory event with its corresponding time frame within the video. The discrepancy reveals a critical challenge in the Temporal Attribute task: models' inability to precisely capture the timing and duration of sound events, even when the total video duration is known.

This case highlights the need for better temporal grounding and understanding in multimodal models. Accurate identification of the specific timing of audio cues is crucial for tasks that rely on precise temporal reasoning.

\begin{figure}[!t]
    \centering
    \includegraphics[width=\linewidth]{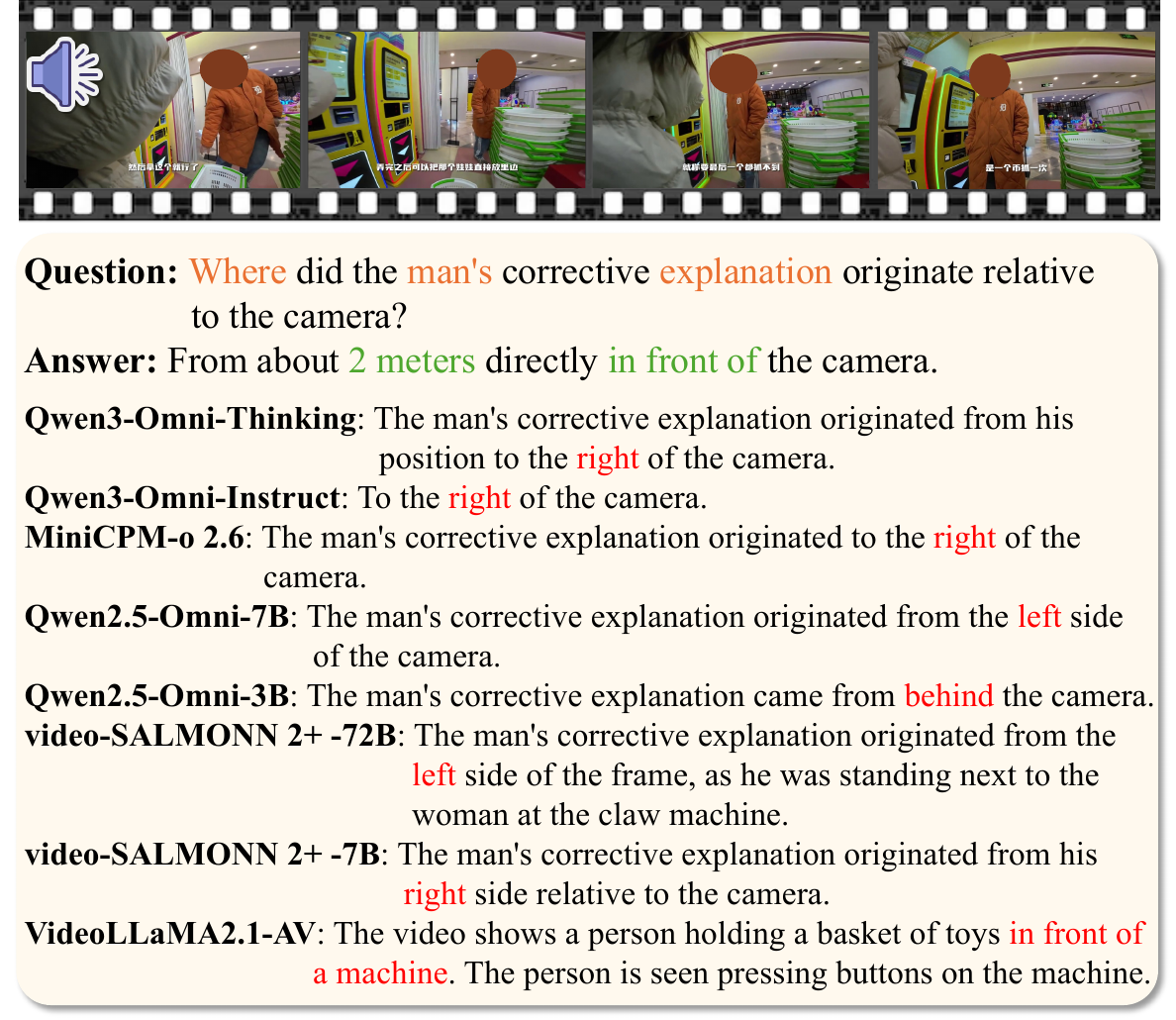}
    \caption{MLLMs fail to localize sound sources under dynamic egocentric motion.}
    \label{fig:saptial}
\end{figure}

\paragraph{Failure case of Spatial Location.}
Fig.~\ref{fig:saptial} illustrates a failure case in the spatial location task. The correct answer is that the man’s corrective explanation originates from roughly two meters directly in front of the camera. However, the models produce highly inconsistent spatial predictions, placing the sound source to the right, to the left, or even behind the camera.
A key reason for these errors is that both the camera and the man are moving during the interaction, causing the spatial cues in the audio to shift dynamically across the segment. Current MLLMs are not yet robust to such egocentric motion: they struggle to integrate evolving visual context with continuously changing auditory directionality. As a result, the models fail to establish a stable spatial reference frame and cannot reliably infer where the sound originates.

This case highlights a core limitation revealed by EgoSound: accurate sound localization in first-person videos requires models to handle dynamic viewpoint changes, a capability that remains underdeveloped in existing MLLMs.

\begin{figure}[!t]
    \centering
    \includegraphics[width=\linewidth]{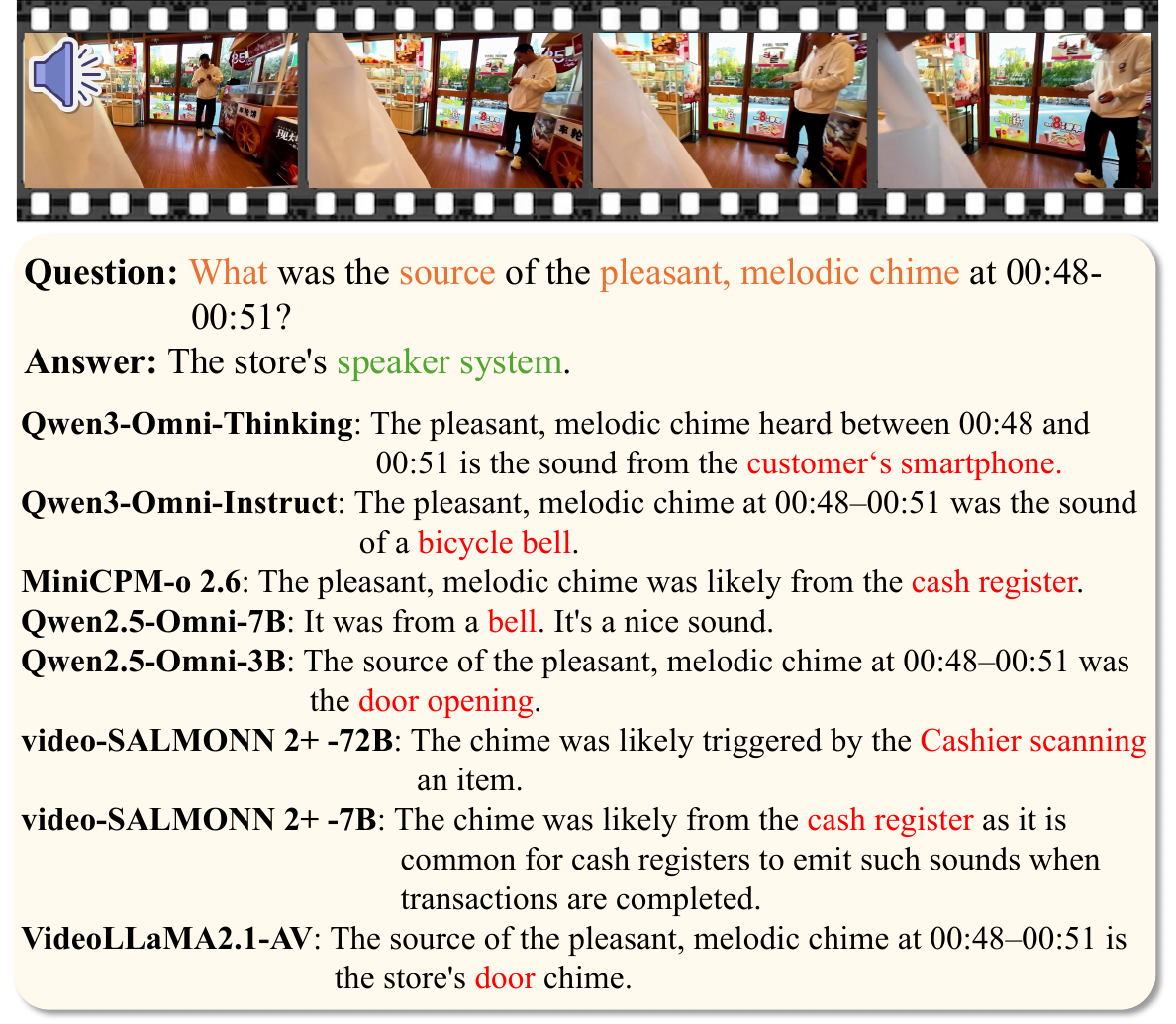}
    \caption{MLLMs fail to identify off-screen sound sources without visual grounding.}
    \label{fig:identification}
\end{figure}

\paragraph{Failure case of Sound Source Identification.}
Fig.~\ref{fig:identification} shows a failure case in the Sound Source Identification task. The correct answer is that the pleasant, melodic chime originates from the store’s speaker system, which is not visible in the video frames. However, most models incorrectly attribute the sound to visually present objects, such as the customer’s smartphone, a bicycle bell, the door opening, or the cash register. A central reason for these errors is that the actual sound source is off-screen, coming from the store’s automated speaker system. Correct identification requires using broader video context—such as noticing that the camera wearer has already moved away from the counter—but many models instead rely only on objects currently visible, leading to systematic misattribution.

This case highlights that MLLMs often struggle when the auditory source is not visually grounded within the current frame, revealing their difficulty in performing sound–context integration and in reasoning beyond immediate visual cues.

\paragraph{Failure case of Inferential Causality.}
\begin{figure}[!t]
    \centering
    \includegraphics[width=\linewidth]{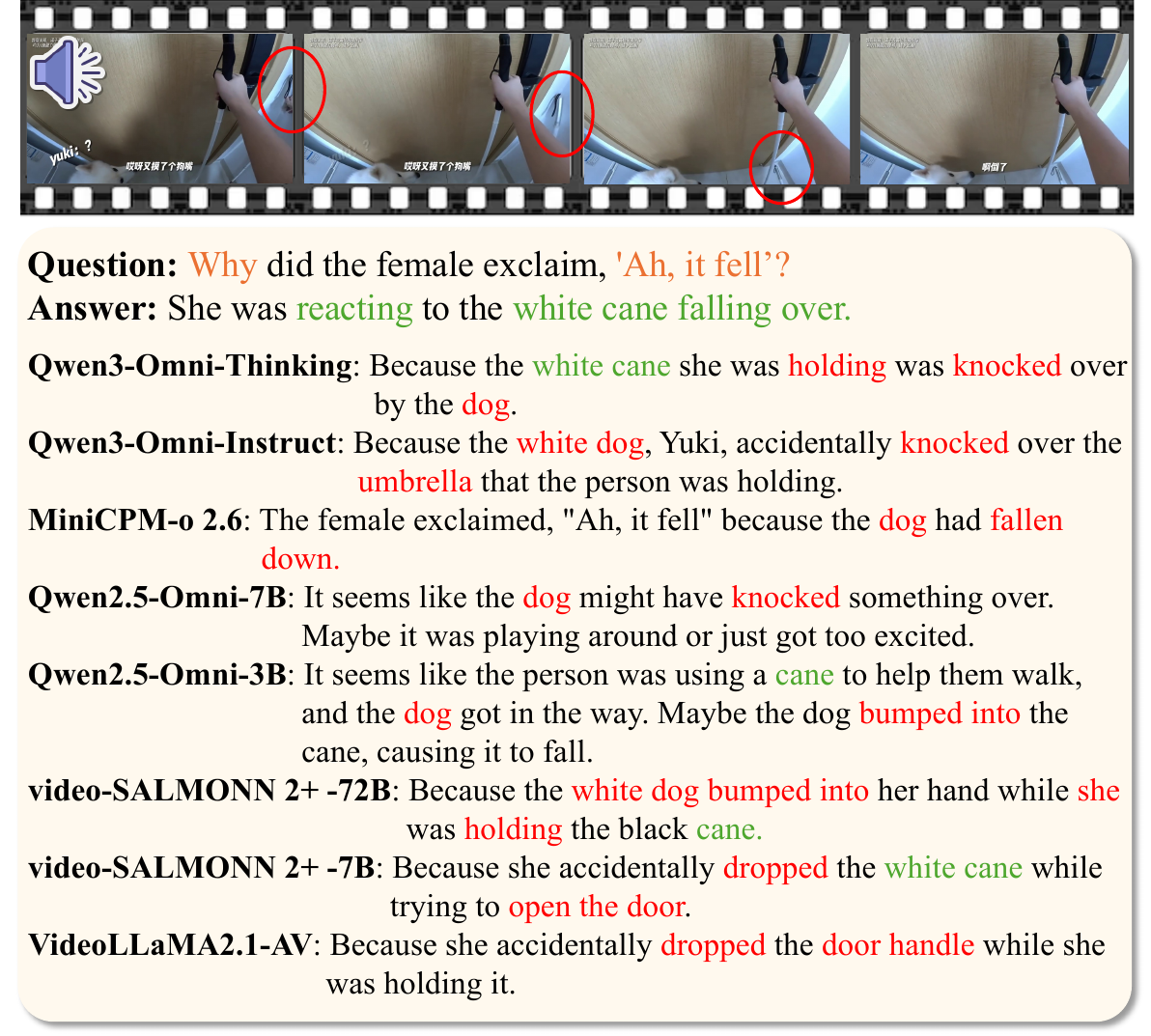}
    \caption{MLLMs fail to infer the causality in egocentric video without prior human interaction information.}
    \label{fig:causality}
\end{figure}
\begin{figure}[!t]
    \centering
    \includegraphics[width=\linewidth]{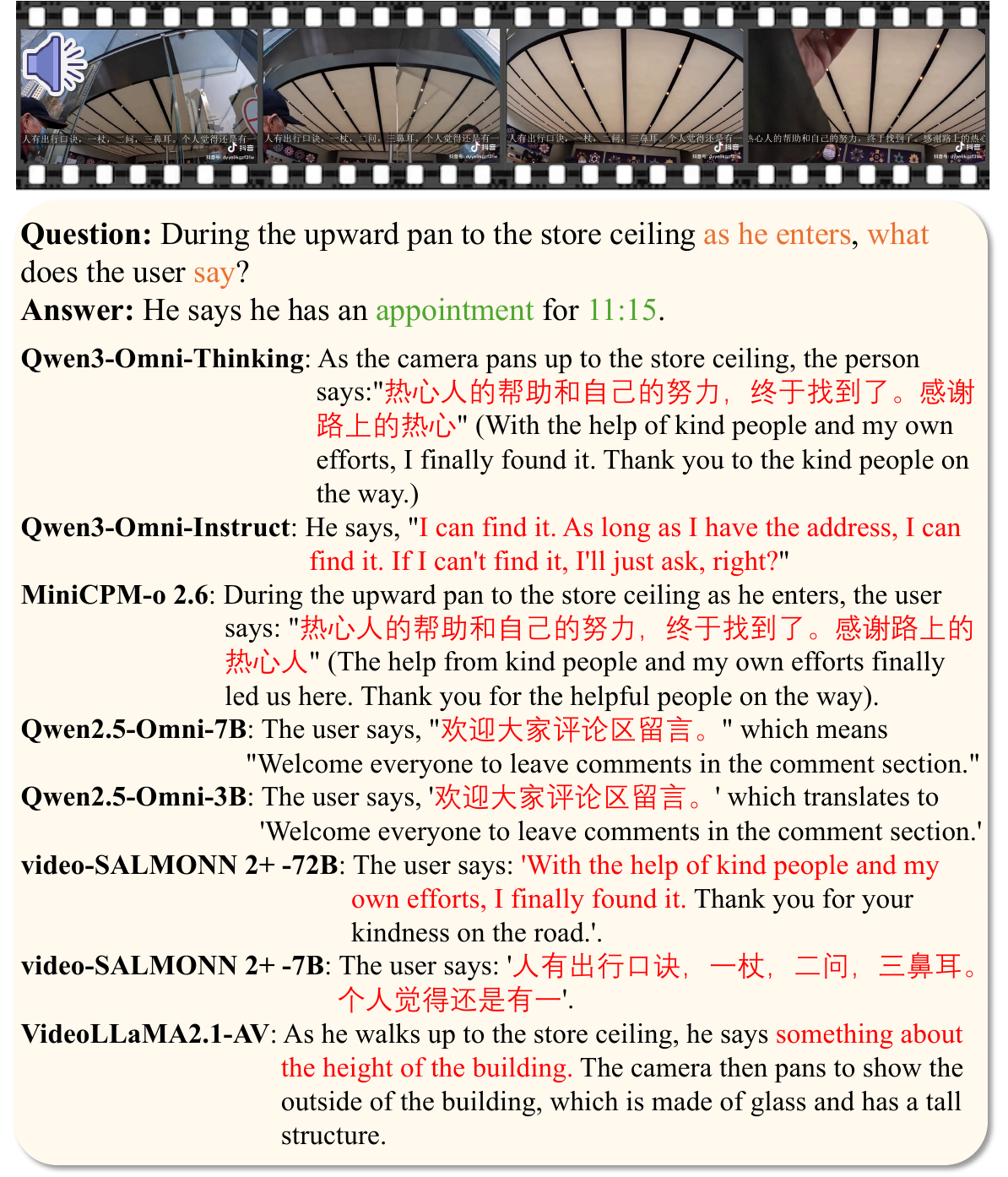}
    \caption{MLLMs fail to temporally align audio with visual cues for cross-modal reasoning.}
    \label{fig:cross}
\end{figure}
Fig.~\ref{fig:causality} reveals a typical failure pattern in current MLLMs. In this scenario, the female exclaims, “Ah, it fell!” because a hand briefly touches the white cane leaning in the corner, after which the cane slides and eventually falls. Although this causal interaction is clearly visible in the egocentric video, none of the evaluated models identified it. Several models instead focus on the nearby white dog and mistakenly attributed the claim to the dog’s actions, while others produced hallucinated explanations entirely unrelated to the scene. Most models did not even recognize that the cane had fallen.

This case shows that, without explicit signals indicating human-object interactions, models often overlook the subtle yet decisive physical events that give rise to specific auditory outcomes. This example therefore highlights the necessity of incorporating interaction-aware annotations into the data curation pipeline: by providing precise temporal grounding of human–object and human–human interaction, such annotations supply the contextual structure needed for models to align visual actions with their corresponding sounds and perform more reliable cross-modal causal reasoning. 

\paragraph{Failure case of Cross-Modal Reasoning.}
Fig.~\ref{fig:cross} presents a failure case in the Cross-Modal Reasoning task. The correct answer is that the user says he has an appointment for 11:15 as the camera pans upward to the store ceiling during entry. However, most models produce entirely unrelated responses—many simply repeat the Chinese subtitles overlaid in the video or hallucinate arbitrary dialogue.

The main cause of these errors is that the models fail to temporally align the visual cue (``as he enters, during the upward pan to the ceiling”) with the corresponding audio segment. Instead of grounding the spoken utterance to the correct visual moment, the models rely on visible on-screen text or generic context, leading to cross-modal mismatches. This case highlights that current MLLMs still struggle to integrate visual timing cues with audio content, revealing a core limitation in fine-grained cross-modal reasoning.

% {   %\clearpage 
%     \small
%     \bibliographystyle{ieeenat_fullname}
%     \bibliography{main}
% }
\end{document}